\documentclass[10pt,onecolumn,letterpaper]{article}
\usepackage{times}
\usepackage{epsfig}
\usepackage{graphicx}
\usepackage{amsmath}
\usepackage{amssymb}
\usepackage{color}
\usepackage{amsthm}
\usepackage[hyphens]{url}
\usepackage{algorithm}
\usepackage{algorithmic}
\usepackage{subfigure}

\makeatletter
\renewcommand{\@thesubfigure}{}

\newcommand{\Figure}[1]{Fig.~{\color{red}#1}}

\newcommand{\MyCaption}[1]{\caption{{#1}}}

\newcommand{\mytextbf}[1]{\vspace{.1cm}\noindent\textbf{{#1}}}

\newcommand{\Table}[1]{Tab.~#1}

\newtheorem{Theorem}{Theorem}

\newtheorem{Lemma}{Lemma}
\newcommand{\LemmaRef}[1]{Lemma.~#1}

\providecommand{\myproofname}{\footnotesize{Proof}}

\begin{document}

\title{Scalable $k$-NN graph construction\thanks{A conference version appeared in~\cite{WangWZTGL12}}}

\author{Jingdong Wang\footnotemark[2]~~~~Jing Wang\footnotemark[3]~~~~Gang Zeng\footnotemark[3]~~~~Zhuowen Tu\footnotemark[2]~~\footnotemark[4]~~~~Rui Gan\footnotemark[3]~~~~Shipeng Li\footnotemark[2]\\
\footnotemark[2]~Microsoft Research Asia~~~~~~\footnotemark[3]~Peking University~~~~~~\\
\footnotemark[4]~Lab of Neuro Imaging and Department of Computer Science, UCLA\\
{\tt \small cis.wangjing@pku.edu.cn}~~~
{\tt \small \{g.zeng, rui\_gan\}@ieee.org}~~~\\
{\tt\small \{jingdw, zhuowent, spli\}@microsoft.com}
}
\date{~}
\maketitle

\begin{abstract}
The $k$-NN graph has played a central role in increasingly popular data-driven techniques
for various learning and vision tasks;
yet, finding an efficient and effective way to construct $k$-NN graphs remains a challenge,
especially for large-scale high-dimensional data.
In this paper, we propose a new approach
to construct approximate $k$-NN graphs with emphasis in: efficiency and accuracy.
We hierarchically and randomly divide the data points into subsets
and build an exact neighborhood graph over each subset,
achieving a base approximate neighborhood graph;
we then repeat this process for several times to generate multiple neighborhood graphs,
which are combined
to yield a more accurate approximate neighborhood graph.
Furthermore, we propose a neighborhood propagation scheme
to further enhance the accuracy.
We show both theoretical and empirical accuracy and efficiency of our approach
to $k$-NN graph construction
and demonstrate significant speed-up in dealing with large scale visual data.
\end{abstract}

\section{Introduction}
The fields of machine learning, computer vision, data mining, bioinformatics,
and internet search have witnessed a great success of applying data-driven techniques,
in which neighborhood graphs are widely adopted.
Some examples include
image and object organization~\cite{HeathGOAG10, PhilbinSZ11, PhilbinZ08},
object retrieval~\cite{JegouSHV10, QinGBQG11},
face synthesis~\cite{Kemelmacher-ShlizermanSGS11},
shape retrieval and approximate nearest neighbor search~\cite{AryaM93, SebastianK02},
manifold learning and dimension reduction~\cite{BelkinN03, RoweisS00, TenenbaumSL00},
and other machine learning tasks~\cite{LiuHC10, MaierHL07, WangWZSQ09, ZhouWGBS03, Zhu05}.

Two types of neighborhood graphs are often used:
$k$-nearest-neighbor ($k$-NN) graphs
(a node is connected to its $k$ nearest neighbors)
and $\epsilon$-nearest-neighbor ($\epsilon$-NN) graphs
(two nodes are connected if their distance is within $\epsilon$).
The $\epsilon$-NN graph is geometrically motivated~\cite{BentleySW77},
but it easily results in disconnected components~\cite{BelkinN03}.
Hence, it is not suitable in many situations~\cite{ChenFS09}.
In contrast,
$k$-NN graphs have been shown to be especially useful in practice~\cite{ChenFS09}.
Therefore, this paper focuses on the problem of constructing $k$-NN graphs.

A naive solution to neighborhood graph construction
is exhaustively comparing all pairs of points,
which takes $\Theta(n^2d)$ time
with $n$ denoting the number of data points
and $d$ denoting the dimensionality.
This is prohibitively slow
and unsuitable for large-scale applications.
Early research efforts have been conducted
to construct exact $k$-NN graphs~\cite{Bentley80, Clarkson83, ParedesCFN06, Vaidya89}.
However, the time complexity of those methods either
grows exponentially with respect to the dimensionality,
or grows super-linearly with respect to the number,
which makes them impractical
for large scale and high-dimensional problems.
Nowadays most research efforts have been turned to
approximate neighborhood graph construction.

A straightforward solution is to
apply approximate nearest neighbor search methods
to construct neighborhood graphs.
One can first build an indexing structure
to organize the data points,
and then regard each data point as a query
and find approximate NNs by searching the structure.
An example approach uses locality sensitive hashing
to help construct neighborhood graphs~\cite{UnoST09}.
However, neighborhood graph construction is
generally simpler than nearest neighbor search
because any NN search algorithm can solve neighborhood graph construction,
but not vice versa~\cite{ChenFS09, UnoST09}.
In other words, neighborhood graph construction only
cares about the existing data points,
while NN search has to consider out-of-sample points.

Approximate neighborhood graph construction methods
have been proposed by following the divide-and-conquer methodology~\cite{Bentley80}.
The process consists of two stages:
recursively divide the data set into small subsets
and merge the neighborhood graphs from subsets.
Using the overlapped divisions~\cite{ChenFS09}
or the third subset~\cite{UnoST09},
the neighborhood graphs from all the subsets
are merged together to get an approximate neighborhood graph.
Due to the overlapped divisions,
both the two approaches suffer from the high time complexity~\cite{ChenFS09}.
The time cost relies much on the overlapping ratio,
and tends to be quadratic with respect to the number of points
for highly accurate neighborhood graphs.

In this paper, we propose a new approach to construct approximate
$k$-NN graphs with emphasis in efficiency and accuracy
and justify our approach in theory and empirical experiments
with large scale vision data.
We first present a multiple random divide-and-conquer approach
to construct an approximate neighborhood graph.
We randomly and hierarchically partition the data points
into subsets
so that the neighboring points have a high probability
to lie in the same subset,
and connect each point with its nearest neighbors
within each subset
to get a base approximate neighborhood graph.
This random partition process is repeated several times
to increase the chance that neighboring points
are connected in at least one random partition.
The multiple random partitions can be viewed
as a way to exploit overlaps
that are also utilized in~\cite{ChenFS09, VirmajokiF04},
but differently our approach is more efficient
and the time cost grows only linearly
with respect to the number of random divisions.

Furthermore,
we propose a neighborhood propagation scheme,
i.e., propagating the local approximate neighborhoods
to the wider area,
to achieve a more accurate neighborhood graph
in a higher speed.
We observe that
after several repetitions of random partitions
most of the neighboring relationships generated
by the new random partition
have already appeared in the previous random partitions
and that
the discovered approximate neighborhoods of true neighboring points
will instead have relatively large overlaps.
The two points suggest to propagate the local neighborhood
to a wider range,
expecting a higher speed in connecting neighboring points.
Experimental results show that
the multiple division scheme is superior over
existing neighborhood construction algorithms
in terms of speed and accuracy,
neighborhood propagation can further improve the performance a lot
and the improvement is more significant
when requiring larger neighborhoods.

\begin{figure*}[t]
\label{NeighborIllustration}
\centering
\subfigure[(a)]
{\label{NeighborIllustration1}
\includegraphics[width = 0.6\linewidth, clip]{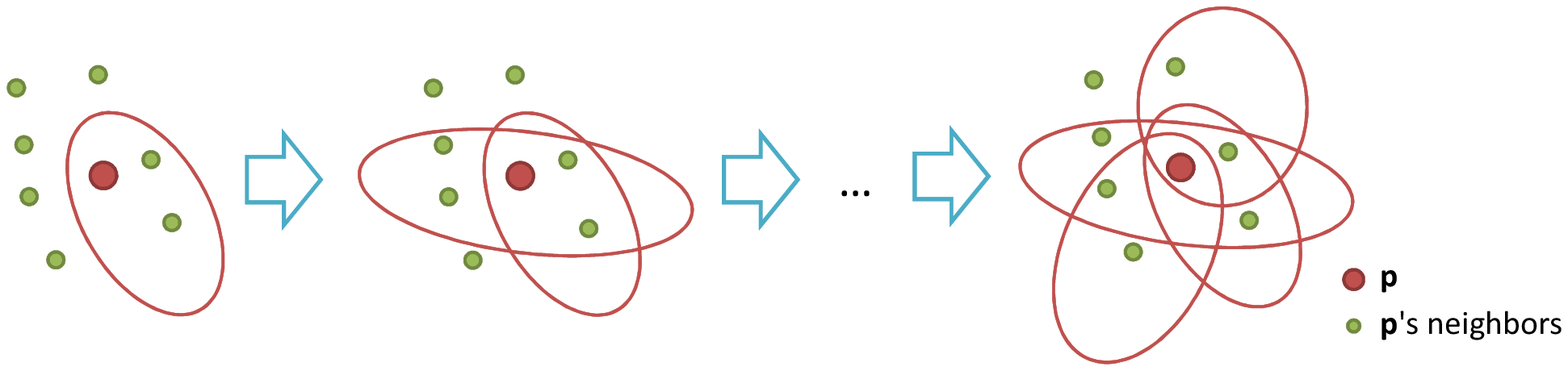}}
\hspace{1.5cm}
\subfigure[(b)]
{\label{NeighborIllustration2}
\includegraphics[width = 0.16\linewidth, clip]{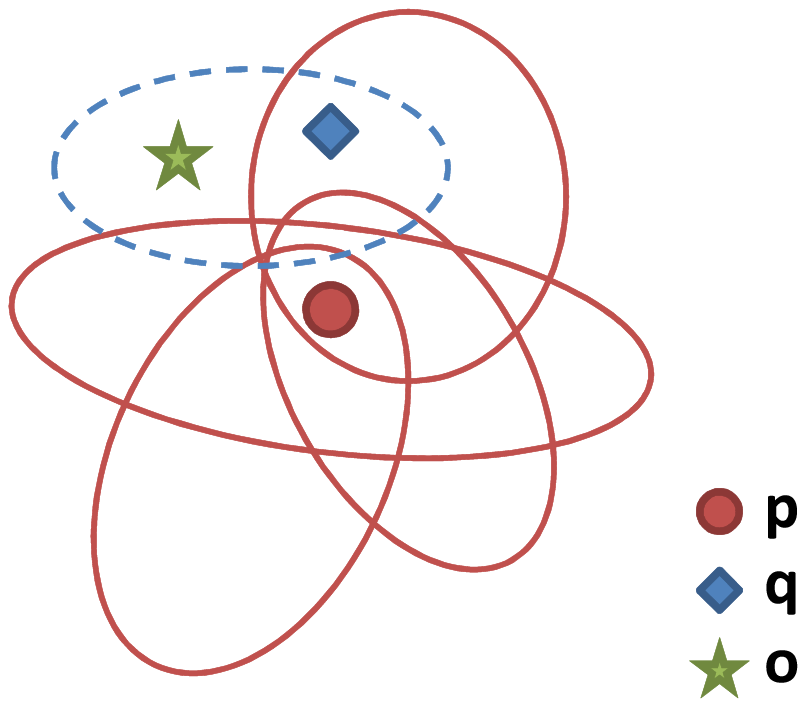}}
\caption{(a) As the number of random divisions increases, more neighbors
of $\mathbf{p}$ are covered.
(b) Illustration of neighborhood propagation from $\mathbf{p}$
to $\mathbf{o}$ by accessing the neighborhood of $\mathbf{q}$}.
\vspace{-0.6cm}
\end{figure*}

\section{Related work}
The construction of exact $k$-NN graphs
has been extensively studied in the literature
and a number of algorithms were proposed
to avoid the high complexity $O(n^2)$.
In~\cite{Bentley80}, a divide-and-conquer method
taking $O(n \log ^{d-1}n)$ time was presented,
while an algorithm with expected $O(c^d \log n)$ time (for some constant $c$)
was introduced in~\cite{Clarkson83}
and a worst-case $O((c'd)^dn \log n)$ time (for some constant $c'$)
algorithm was proposed in~\cite{Vaidya89}.
The time complexity of these methods grows
exponentially with data dimension $d$,
which makes them quite inapplicable in high-dimensional problems.
Recently,~\cite{ParedesCFN06} proposed a method,
which empirically requires $O(n^{1.27})$ distance calculations
in low-dimensional cases and $O(n^{1.90})$ calculations
in high-dimensional cases.
This algorithm works well on low-dimensional data,
but becomes inefficient in high-dimensional cases.
In spite of a rich previous literature,
no efficient algorithm for high-dimensional exact
$k$-NN graphs has been proposed.
Thus, research efforts are moved
to approximate neighborhood graph construction.

A straightforward solution of constructing
an approximate $k$-NN graph is to apply
a nearest neighbor search algorithm,
in which an indexing structure
is usually first made
to organize the data points
and a search method based on the structure is adopted
to handle the upcoming queries.
Representative examples of search methods
are partition-tree-based methods
such as kd-trees~\cite{Bentley75, FriedmanBF77, JiaWZZH10, JonesOR11},
and random projection trees (rp-tree)~\cite{DasguptaF08},
and hashing based methods such as
locality sensitive hashing~\cite{DatarIIM04,UnoST09}.
However, as pointed in~\cite{ChenFS09},
the existing search methods generally suffer from
an unfavorable trade-off between the complexity
of the indexing structure and the accuracy of the search.
Moveover, the search methods generally ignore the fact
that in graph construction,
each query must be one of the data points.
As a consequence, unnecessary efforts are put on
giving a good result for general queries,
which makes them not so efficient comparing
to other algorithms which focus only on graph construction.

Some approximate neighborhood graph construction methods
were proposed recently
by following the divide-and-conquer methodology~\cite{Bentley80}.
The approach in~\cite{ChenFS09} divides the data points
into two or three overlapped subsets
with a predefined overlapping ratio,
and unites the subgraphs constructed from subsets together,
which is followed by a refinement step,
inspecting the neighbors of the neighbors
(or the second-order-neighbors).
Our neighborhood propagation is related to this refinement step,
but very different
because the refinement
does not discriminate the points
in the second-order neighborhood
and may check some distant points.
In contrast,
our neighborhood propagation inspects the points
within higher-order neighborhood
in the best-first order,
which is more reasonable.
The approach in~\cite{VirmajokiF04} divides the data into two non-overlapped subsets
and additionally samples another subset which overlaps with both the above two subsets,
and finally merges the three graphs together.
The accuracy of both methods rely much on
the overlapping ratio,
but the time cost grows almost exponentially with the ratio,
which makes them difficult to balance
the efficiency and the accuracy especially in large scale problems.
The approach proposed in~\cite{JonesOR11} is very related to our approach,
but is clearly different from ours
as it adopts randomly rotated kd-trees to perform an approximate search
and then performs a second-order neighborhood expansion.

There are some other approaches of neighborhood graph construction.
\cite{ConnorK10} proposed a parallel fast algorithm using Morton ordering,
but the method works well only on low-dimensional data.
\cite{HacidY07} presented an incremental way of building neighborhood graphs,
but the algorithm is mainly for relative neighborhood graph
and is inefficient in large scale problems.
\cite{MaierHL07} gave an extensive theoretical analysis
on how to choose a proper $k$ in $k$-NN graphs
for a better performance in real applications,
which is a problem different from our approach.

The multiple random division scheme
has been exploited in other problems.
Random kd-trees are constructed in~\cite{Silpa-AnanH08}
to boost the indexing efficiency.
Random forest~\cite{Breiman01}
is developed
to build an ensemble classifier
that consists of many decision trees
to improve the classification performance.
Differently,
this paper proposes to adopt this multiple random division technique
to build a neighborhood graph.

\section{Approach}
Given a set of data points
$\mathcal{X} = \{\mathbf{x}_1,~\mathbf{x}_2,~\cdots, \mathbf{x}_n\}$
with $\mathbf{x}_i \in \mathbb{R}^d$,
the goal is to build
a $k$-nearest-neighbor graph $G = (V,E)$.
The problem is formally defined as follows.

\newtheorem{definition}{Definition}
\label{def:knn}
\begin{definition}[$k$-NN Graph]
$G$ is a directed graph,
where $V = \mathcal{X}$,
and $\langle \mathbf{x}_i, \mathbf{x}_j \rangle \in E$
if and only if $\rho(\mathbf{x}_i,\mathbf{x}_j)$
is among the $k$ smallest elements of the set
$\{\rho(\mathbf{x}_i,\mathbf{x}_l)|l=1,\dots,i-1,i+1,\dots,n\}$,
where $\rho$ is a metric such as Euclidean distance and cosine distance.
\end{definition}

We first present a multiple random divide-and-conquer approach
to build base approximate neighborhood graphs
and then ensemble them together
to achieve an approximate neighborhood graph.
Then we introduce a neighborhood propagation technique
to propagate the local neighborhoods to a wider range
in order to achieve a more accurate neighborhood graph
in a higher speed.

\subsection{Multiple random divide-and-conquer}
\label{sec:mrd}

A base approximate neighborhood graph is an unconnected graph,
in which each subgraph corresponds to a group of possibly neighboring points.
In other words, a base approximate neighborhood graph
corresponds to a partitioning of the data points.
We adopt the divide-and-conquer methodology
to recursively partition the points into small subsets,
forming a random partition tree.
To make nearby points lie in the same subset,
we can use hyperplanes or hyperspheres
to partition the data set.
Specifically,
we divide the point set $\mathcal{X}$
into two nonoverlapped subsets,
$\mathcal{X}_l$ and $\mathcal{X}_r$,
so that $\mathcal{X}_l \cup \mathcal{X}_r = \mathcal{X}$
and $\mathcal{X}_l \cap \mathcal{X}_r = \emptyset$,
and recursively conduct the division process
on the subsets
until the cardinality of a subset is
smaller than a fixed value $g$.
Then a brute-force manner is adopted
to build a neighborhood graph (subgraph)
for each subset of points.
Different from the division with the overlapping in~\cite{ChenFS09, UnoST09},
this process is very efficient as the
partitioning process takes $O(nd \log n)$ and building subgraphs only takes $O(ndg)$.

A single random division yields
a base approximate neighborhood graph
containing a serial of isolated subgraphs,
and it is unable to connect a point
with its neighboring points
lying in different subgraphs.
Thus,
we propose to exploit multiple random divisions
to find more neighbors for each point.
Considering a point $\mathbf{p}$,
each random division can be interpreted
as enumerating a set of neighboring points around $\mathbf{p}$.
As illustrated in~\Figure{\ref{NeighborIllustration1}},
the neighborhood of $\mathbf{p}$
identified in each division is
represented by a red ellipse.
We denote the set of points in the neighborhood of $\mathbf{p}$
in the $m$-th division by $\mathcal{N}^m_{\mathbf{p}}$,
and the union of multiple sets of neighboring points
is then written as $\tilde{\mathcal{N}}^m_{\mathbf{p}} = \cup_{i \leqslant m} \mathcal{N}^i_{\mathbf{p}}$.
By increasing the number of random divisions,
the union $\tilde{\mathcal{N}}^m_{\mathbf{p}}$
will cover more true neighbors of $\mathbf{p}$,
which are represented by small green points in~\Figure{\ref{NeighborIllustration1}},
so that the quality of the combined neighborhood graph is improved.
We denote the base approximate neighborhood graphs resulted from each division
by $\{G_1, \cdots, G_M\}$,
and the adjacent list by $Adj^m[\mathbf{x}_i]$ in $G_m$ for point $\mathbf{x}_i$.
The combination is achieved efficiently by
uniting its adjacent lists $\{Adj^m[\mathbf{x}_i]\}_{m=1}^M$ together
and retaining $k$ nearest neighbors.

\begin{figure}[t]
\centering
\subfigure[(a)]
{
\label{division1}
\includegraphics[width = 0.25\linewidth, clip]{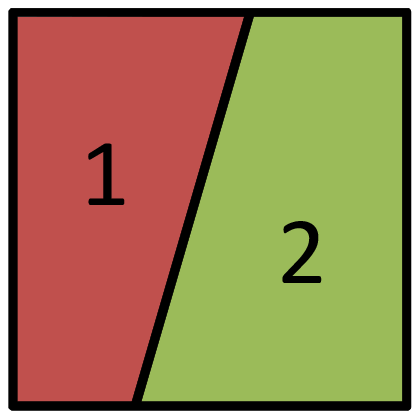}
}
\hspace{.2cm}
\subfigure[(b)]
{
\label{division2}
\includegraphics[width = 0.25\linewidth, clip]{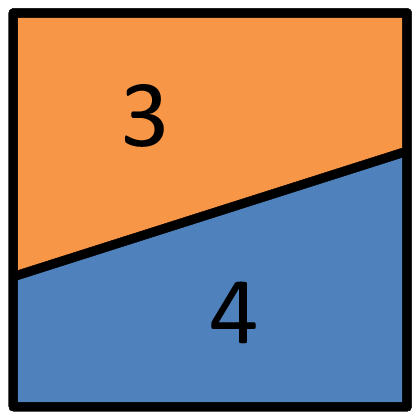}
}
\hspace{.2cm}
\subfigure[(c)]
{
\label{overlap}
\includegraphics[width = 0.25\linewidth, clip]{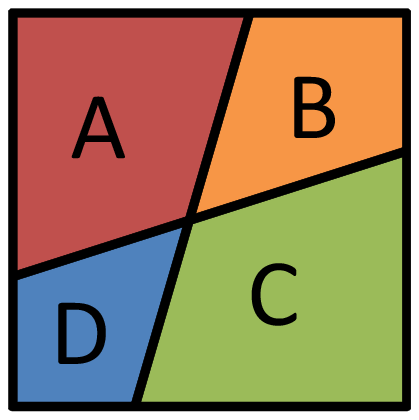}
}
\centering
\caption{Overlaps of subsets in different divisions
served as bridges to connect isolated subgraphs.
}
\label{OverlapIllustration}
\end{figure}

In essence, uniting the neighborhood subgraphs
exploits the overlaps
among the subgraphs from multiple random divisions.
Let's consider a situation of two random divisions
illustrated in~\Figure{\ref{OverlapIllustration}}.
In~\Figure{\ref{division1}} and~\Figure{\ref{division2}},
the first division yields
two isolated subsets $S_1$ and $S_2$,
and the second division yields two isolated subsets
$S_3$ and $S_4$.
In~\Figure{\ref{overlap}},
one can see that
$S_3$ overlaps with $S_1$ over $S_A$
and $S_3$ overlaps with $S_2$ over $S_B$.
This implies that $S_3$ serves as a bridge
to connect isolated subgraphs
constructed from $S_1$ and $S_2$.
$S_4$ also serves as the same role.
In turn, $S_1$ and $S_2$
can also be regarded as the same roles
to connect subgraphs over $S_3$ and $S_4$.
In the implementation,
the data points are divided into many parts
for many times
so that there are sufficient overlaps
to make better connections among subgraphs.

\begin{figure}[t]
\centering
\includegraphics[width = 0.7\linewidth]{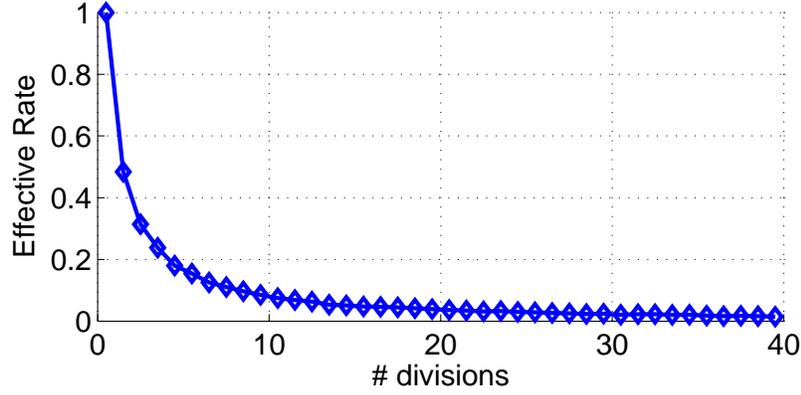}
\caption{Effective rate drops as the increase of the number of random divisions
}.
\label{fig:EffectiveRate}\vspace{-.2cm}
\end{figure}

\subsection{Neighborhood propagation}
\label{sec:neighborhoodpropagation}

Let's consider the point $\mathbf{p}$ again.
As discussed before,
by increasing the number of random divisions,
the union $\tilde{\mathcal{N}}^m_{\mathbf{p}}$
becomes larger and
covers more true neighbors of $\mathbf{p}$.
Although this increases the accuracy of the neighborhood graph,
it also makes the contribution of a new random division
$\tilde{\mathcal{N}}^m_{\mathbf{p}} - \tilde{\mathcal{N}}^{m-1}_{\mathbf{p}}$ smaller.
In other words, the progress
toward the true neighborhood graph
becomes slower when $m$ becomes larger.
This is validated by the experimental result
shown in~\Figure{\ref{fig:EffectiveRate}}.
The effective rate for the $m$-th division is defined
as $r_m = \frac{\Sigma_{\mathbf{p}}| \tilde{\mathcal{N}}^m_{\mathbf{p}} -
\tilde{\mathcal{N}}^{m-1}_{\mathbf{p}}|}
{\Sigma_{\mathbf{p}}|\tilde{\mathcal{N}}^m_{\mathbf{p}}|}$,
which indicates the contribution made by the $m$-th division.
On the other hand,
suppose $\mathbf{q}$ is a point in the previously-identified neighborhood of $\mathbf{p}$,
and $\mathbf{p}$ and $\mathbf{q}$
has a common true neighboring point $\mathbf{o}$.
Then with the increase of the number of random divisions,
the probability that $\mathbf{o}$ is identified as
a neighboring point of $\mathbf{p}$ or $\mathbf{q}$
increases,
i.e., $P(\mathbf{o} \in \mathcal{N}_p~or~\mathbf{o} \in \mathcal{N}_q)$
becomes larger.
This suggests a way
to find the neighboring point $\mathbf{o}$ for $\mathbf{p}$ ($\mathbf{q}$),
by accessing its neighbor $\mathbf{q}$ ($\mathbf{p}$)
and expanding the neighborhood of $\mathbf{q}$ ($\mathbf{p}$).
In the situation illustrated in \Figure{\ref{NeighborIllustration2}},
$\mathbf{o}$ has not been identified as $\mathbf{p}$'s neighbor,
but it has been identified as $\mathbf{q}$'s neighbor
through the subset denoted by the blue dashed ellipse.
Consequently, a neighborhood propagation path
from $\mathbf{p}$ to $\mathbf{o}$ through $\mathbf{q}$
is accessible.

In light of the above analysis,
we present a neighborhood propagation scheme.
For each point $\mathbf{p}$,
we access its previously-identified neighborhood
and conduct more accesses gradually
by propagating the neighborhoods
in a best-first manner.
Specifically,
we first expand $\mathbf{p}$'s neighborhood,
and push all the neighbors into a priority queue,
in which the point the nearest to $\mathbf{p}$ will be positioned
at the top.
Then we iteratively pop the top point from the queue
and push all its unvisited neighbors into the queue.
The best-first strategy
makes true neighbors
be first discovered
with higher probability.
The propagation process stops
when the queue is empty
or the maximum number of visited points, $T$,
is reached.
During the process,
all the visited points
are considered as candidate neighbors of $\mathbf{p}$
in which the better ones will replace the current neighbors.
The propagation process
is performed
for all the points.
The process is very efficient
and the cost
is linear with $n$ and $d$.

\subsection{Analysis}
In the following,
we present theoretic analysis to show
why random divisions and neighborhood propagation work well
and complexity analysis of our approach.
The detailed proofs
can be found in~Section~\ref{sec:appendix}.

\begin{Lemma}
\label{lemma1}
Suppose a random hyperplane partitions the data points
so that a certain point $\mathbf{x}_i$
and one of its true neighbors $\mathbf{x}_j$
have the probability $P_{ij}$ to be in the same subset.
Then with a single random partition tree,
$\mathbf{x}_j$ is discovered as $\mathbf{x}_i$'s neighbor
with the probability $P_{ij}^h$,
where $h$ is the depth of the tree.
With $L$ random partitions,
$\mathbf{x}_j$ is discovered as $\mathbf{x}_i$'s neighbor
with the probability $1 - (1 - P_{ij}^h)^L$.
\end{Lemma}

\begin{Lemma}
\label{lemma2}
The probability that
the neighboring relationship between $\mathbf{x}_i$ and $\mathbf{x}_j$ is discovered
by the $L$-th random partition tree
but not discovered by the previous $L-1$ random partition trees
is $ P_{r_L} = (1 - P_{ij}^h)^{L-1} P_{ij}^h$.
\end{Lemma}
This lemma
indicates that the true NN points newly found in the $L$-th partition tree
become fewer when $L$ increases
and presents a theoretic justification of
\Figure{\ref{fig:EffectiveRate}}.

\begin{Lemma}
\label{lemma3}
Considering two neighboring points $\mathbf{x}_i$ and $\mathbf{x}_j$
having the same neighboring point $\mathbf{x}_n$,
after $L-1$ partition trees,
the probability that $\mathbf{x}_j$ can be discovered as the neighbor of $\mathbf{x}_i$
through $\mathbf{x}_n$
is $ P_{p_L} = (1 - (1- P_{in}^h)^{L-1})(1 - (1- P_{jn}^h)^{L-1})$.
\end{Lemma}

It can be easily seen that
$P_{p_L}$ becomes larger
when $L$ increases.
This property can be generalized to the case
that $\mathbf{x}_j$ is discovered as the neighbor of $\mathbf{x}_i$
through $t$ intermediate points (a longer neighborhood propagation path)
and the probability is $ P_{p_L} = \prod_{k=0}^{t}(1 - (1- P_{n_kn_{k+1}}^h)^{L-1})$,
with $n_0 = i$ and $n_{t+1} = j$.

Let's compare the probabilities
of discovering a new true nearest neighbor
from a partition tree and neighborhood propagation.
Under the condition that $\mathbf{x}_i$ and $\mathbf{x}_j$
are not discovered as neighbors in previous $L-1$ trees,
we have conclusions:
(1) The probability $P_{ij}^h$ to discover the neighboring relationship
in the next tree stay the same as $L$ grows;
(2) The probability $P_{pL}$ to discover the relationship
by propagation keeps increasing;
(3) The probability with the next tree will be smaller
than that with propagation when $L$ reaches one constant.
This shows that
neighborhood propagation can speed up neighborhood discovery.
It should be noted that
neighborhood propagation is even more advantageous
because the above analysis does not cover all the cases,
for instance,
when $\mathbf{x}_i$ and $\mathbf{x}_j$ have
more same neighbors.
In summary, we can have the following theorem.

\begin{Theorem}
\label{theorem1}
Suppose $P = \min\{ P_{ij} | \langle \mathbf{x}_i,\mathbf{x}_j \rangle \in E(G) \}$,
where $G$ is the exact $k$-NN graph.
With $L$ random divisions and a first-order neighborhood propagation,
a true $k$-NN point, $\mathbf{x}_j$,
of $\mathbf{x}_i$ is discovered
with at least the probability $1 - (1 - P^h)^{2L}(2-(1-P^h)^L)$
under the assumption
that $\mathbf{x}_i$ and $\mathbf{x}_j$
have at least one same neighboring point.
\end{Theorem}

The following discusses the time complexity.
Our approach takes $O(Mdn \log n)$ time
in multiple random divisions,
where $M$ denotes the number of divisions,
and $O(Tdn \log T)$ time in neighborhood propagation,
where $T$ denotes the maximum number of visited points.
In a large scale and high-dimensional problem,
$M$ and $T$ are relatively very small
so that the whole complexity of
both the multiple random divisions
and the combined method can be written as $ O(dn \log n) $.
The algorithm presented in~\cite{VirmajokiF04}
is denoted by VirmajokiF04
and its complexity is reported as $ O(d^2n^{1.58} \log n) $.
The two algorithms in~\cite{ChenFS09},
which are named as Glue and Overlap,
take $ O(dn^{1.22} \log n) $ time and $ O(dn^{1.36} \log n) $ time
when the overlapping ratio $ \alpha $ is set as $0.2$
that is suggested in~\cite{ChenFS09}.
By comparison,
the theoretic time complexity of our approach
is smaller that those of other methods.

\subsection{Algorithm details}
\mytextbf{Random division.}
Our implementation chooses the random principal directions
to perform random divisions
to make the diameter of each subset
small enough.
It is theoretically shown in~\cite{VermaKD09} that
the principal-direction-based way
to hierarchically partition the points
can guarantee that the diameters of the subsets
are reduced quickly.
This implies that
the points in the same subset
tend to be nearer to each other.
The principal directions are obtained
by using principal component analysis (PCA).
To generate random principal directions,
rather than computing the principle direction
from the whole subset of points,
we compute the principal direction
over the points randomly sampled
from each subset.
In our implementation,
the principle direction is computed
by the Lanczos algorithm~\cite{Lanczos50}.
Compared with other space partitioning ways,
e.g.,
random projections~\cite{DasguptaF08},
our experiments show
that the principal-direction-based way is more efficient
and effective.

\mytextbf{Speedup.}
In the process
of multiple random divisions
and neighborhood propagation,
the distances between a pair of points
may be computed more than once,
which would cost too much
especially for high-dimensional cases.
To avoid the re-computations,
a hash table is adopted to store
the pairs of points
whose distances have been computed.
When requiring the distance of a pair of points,
we check if their distance has been evaluated
through the hash table.
The time overhead is very low
because the operations over the hash table
cost $O(1)$,
while the re-computation cost is $O(d)$.

Besides,
we introduce a pairwise updating scheme.
This is motivated by the observation
that it is highly possible that
$\mathbf{u}$ is also among the $k$ nearest neighbors of $\mathbf{v}$
if $\mathbf{v}$ is among the $k$ nearest neighbors of $\mathbf{u}$.
When considering $\mathbf{v}$ to update the neighborhood of $\mathbf{u}$,
we also immediately use $\mathbf{u}$ to update
the neighborhood of $\mathbf{v}$.

\section{Experiments}
\mytextbf{Data sets.}
We demonstrate the proposed neighborhood graph construction algorithm
over SIFT features and GIST features.
The SIFT features are collected from
the Caltech 101 data set~\cite{FeiFP04}
and the recognition benchmark images~\cite{NisterS06}.
We extract maximally stable extremal regions (MSERs)
for each image,
and compute a $128$-dimensional SIFT feature for each MSER.
For each image set,
we randomly sample $1000K$ SIFT features
as our data set.

Besides,
we conduct the experiments
on the TinyImage set~\cite{TorralbaFF08}
and the ImageNet data~\cite{DengDSLL009}
to justify our approach.
Similar to~\cite{KulisG09},
we use a global GIST descriptor
to represent each image,
which is a $384$-dimensional vector
describing the texture within localized grid cells.
The dimension of the GIST descriptor is
higher than that of the SIFT feature,
and hence to achieve high accuracy is more difficult and challenging.
We also sample $1000K$ GIST features from each of the two data sets.

\mytextbf{Evaluation scheme.}
We adopt
the accuracy measurement to
evaluate the quality of the approximate graph.
The accuracy of an approximate $k$-NN graph $G'$
(with regard to the exact neighborhood graph $G$)
is defined as $
\operatorname{accuracy}(G') = \frac{ | E(G') \cap E(G) | }{ | E(G) | }$,
where $E(\cdot)$ denotes the set of direct edges in the graph
and $| \cdot |$ denotes the cardinality of the set.
The accuracy is within the range $[0,1]$,
and a higher accuracy means a better graph.
The exact neighborhood graph $G$ is computed
by the brute-force method,
and the running time on four data sets are given in~\Table{\ref{tab:bruteforce}}.

\begin{table}[t]
\MyCaption{The running time of the brute-force method.}
\label{tab:bruteforce}
\centering
{\footnotesize
{
\begin{tabular}{c|cccc}
\hline
& Caltech 101 & Ukbench & Imagenet & TinyImage \\
\hline
Time (min) & 2034 & 2067 & 5737 & 5823 \\
\hline
\end{tabular}}
\vspace{-.2cm}
}
\end{table}

We report the results of our approaches
based on multiple random divide-and-conquer
and the combination of it with subsequently followed
neighborhood propagation.
The recursive division is repeated
till the cardinality of a subset is smaller than $500$.
The neighborhood propagation is triggered
when the effective rate of
the $m$-th random division,
$r_m$, defined in Sec~\ref{sec:neighborhoodpropagation},
is less than a threshold
which is set as $0.05$.
With the help of the hash table,
the effective rate can be easily calculated
and will not affect the efficiency of the algorithm.

By comparison, we present the performances
of the three divide-and-conquer approaches
in~\cite{ChenFS09, VirmajokiF04},
(named ChenFS09Glue, ChenFS09Overlap
and VirmajokiF04, respectively)
and the multisorting algorithm in~\cite{UnoST09}
(named UnoST09)
that leverages locality sensitive hashing.
The results of ChenFS09Glue and ChenFS09Overlap
in~\cite{ChenFS09}
are reported
by running the online implementation
available\footnote{ \textup{http://www.mcs.anl.gov/\~{}jiechen/research/software/knn.tar.gz}}.
We implemented other algorithms
and adjusted the parameters to
make the performance as good as possible.
We also present the performance
of building neighborhood graphs
by searching random kd-trees,
which performs better than other partition trees
as the construction of kd-trees
is very cheap.
All algorithms are run on a $2.66$GHz desktop PC with a single thread.

\begin{table}[t]
\MyCaption{Local label consistency ratio for face organization.}
\label{tab:goodranking}
\centering
{\footnotesize
{
\begin{tabular}{c|ccc}
\hline
& $\mathcal{G}_0$ & $\mathcal{G}_1$ & $\mathcal{G}_2$  \\
\hline
Local label consistency ratio & 0.564 & 0.561 & \textbf{0.689}  \\
\hline
\end{tabular}}
\vspace{-.2cm}
}
\end{table}

\begin{figure*}[ht]
\begin{minipage}[t]{1\linewidth}
\centering
\scriptsize{$ k = 1 $}
\hspace{0.1cm}
\subfigure
{\label{fig:result:101_1nn}
\includegraphics[width=.2\linewidth, clip]{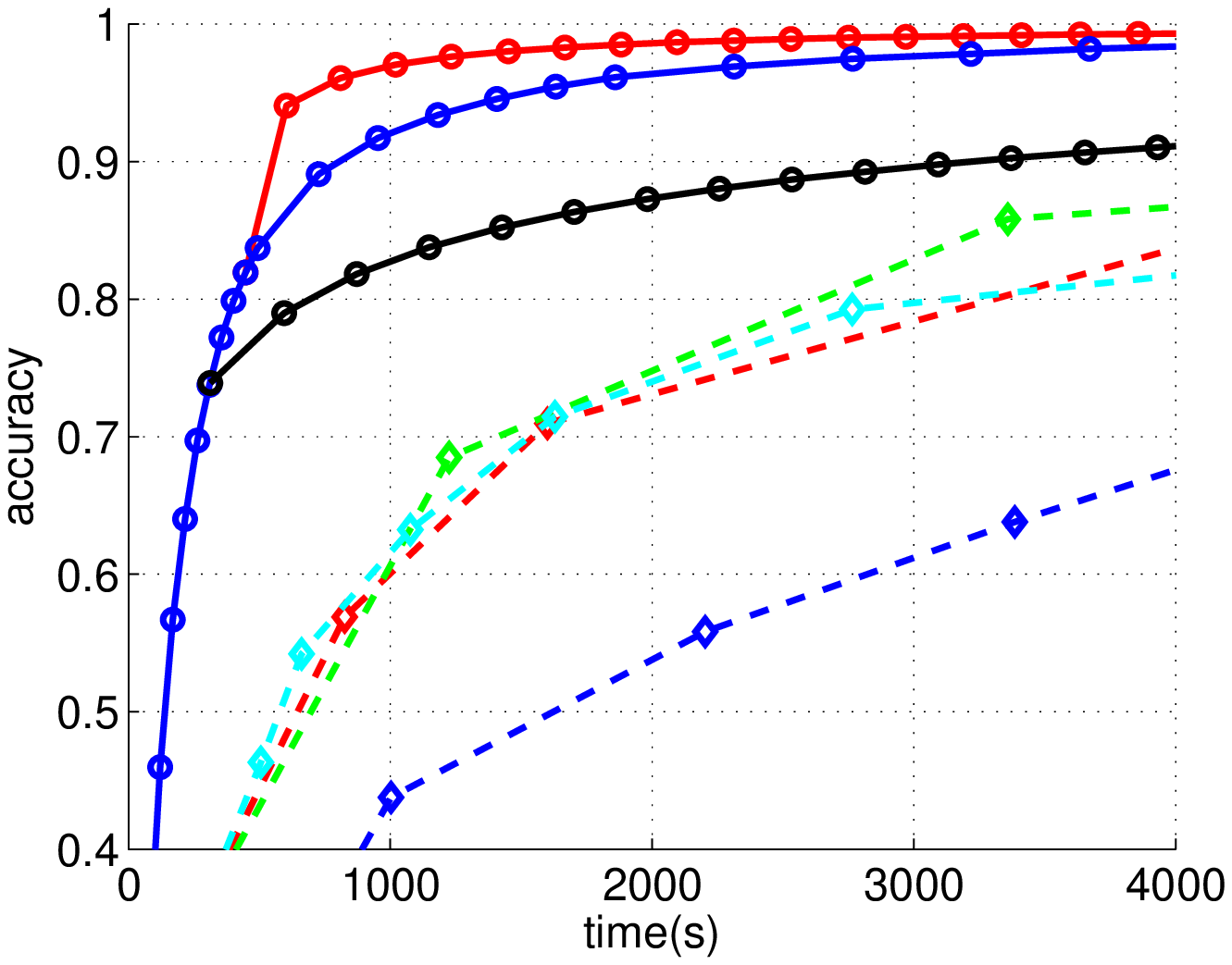}}
\subfigure
{\label{fig:result:uk_1nn}
\includegraphics[width=.2\linewidth, clip]{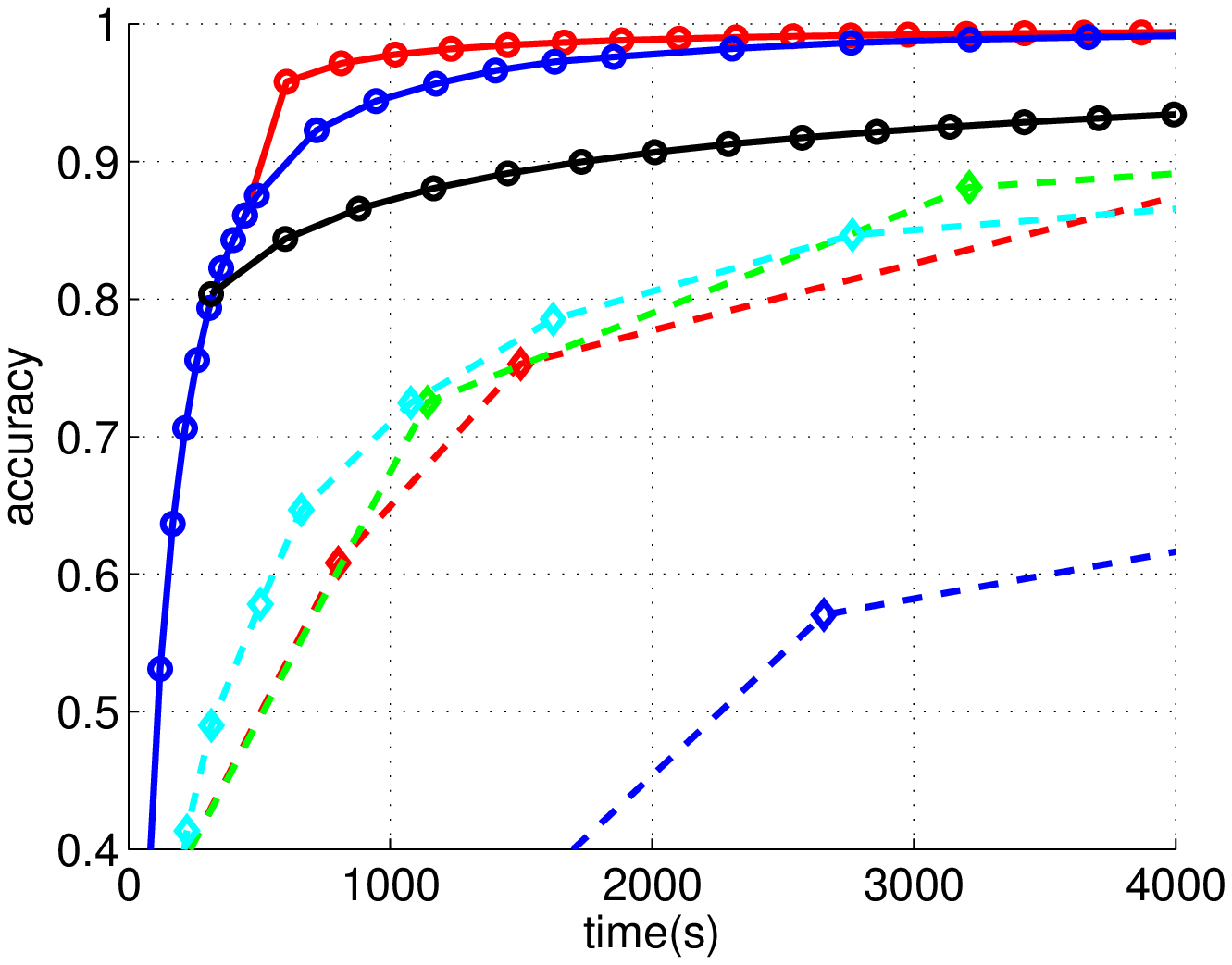}}
\subfigure
{\label{fig:result:imagenet_1nn}
\includegraphics[width=.2\linewidth, clip]{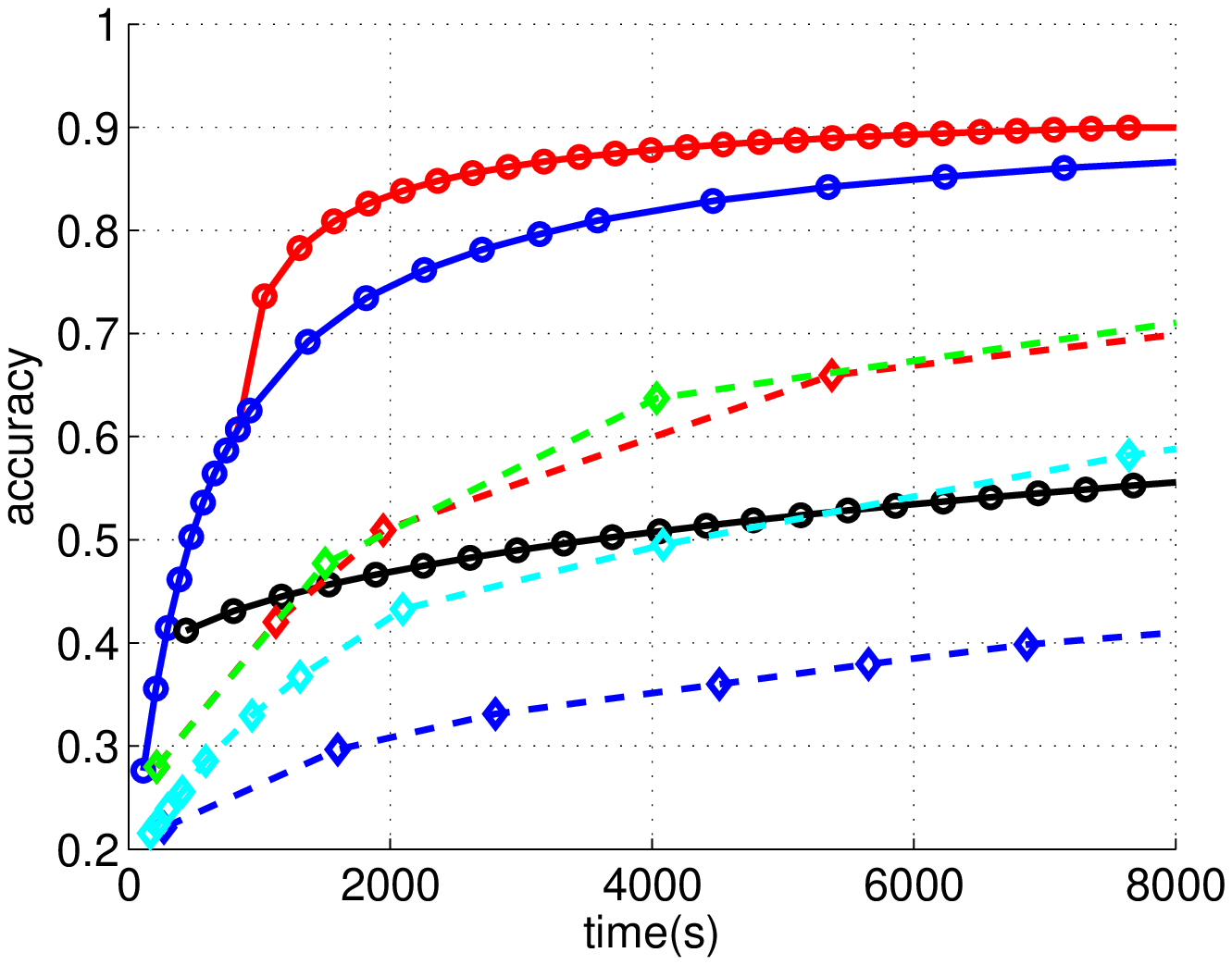}}
\subfigure
{\label{fig:result:tiny_1nn}
\includegraphics[width=.2\linewidth, clip]{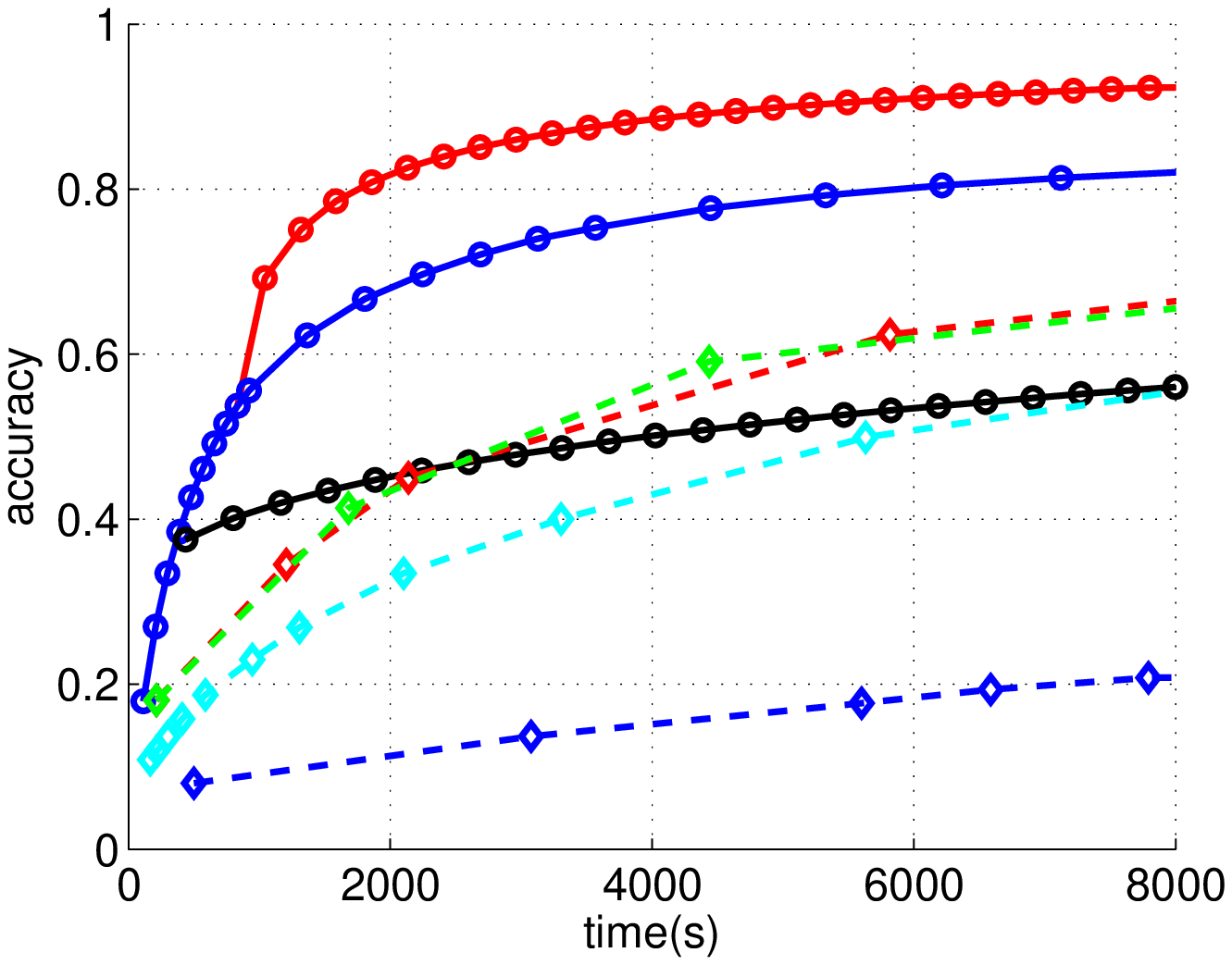}}\\
\scriptsize{$ k = 20 $}
\subfigure
{\label{fig:result:101_20nn}
\includegraphics[width=.2\linewidth, clip]{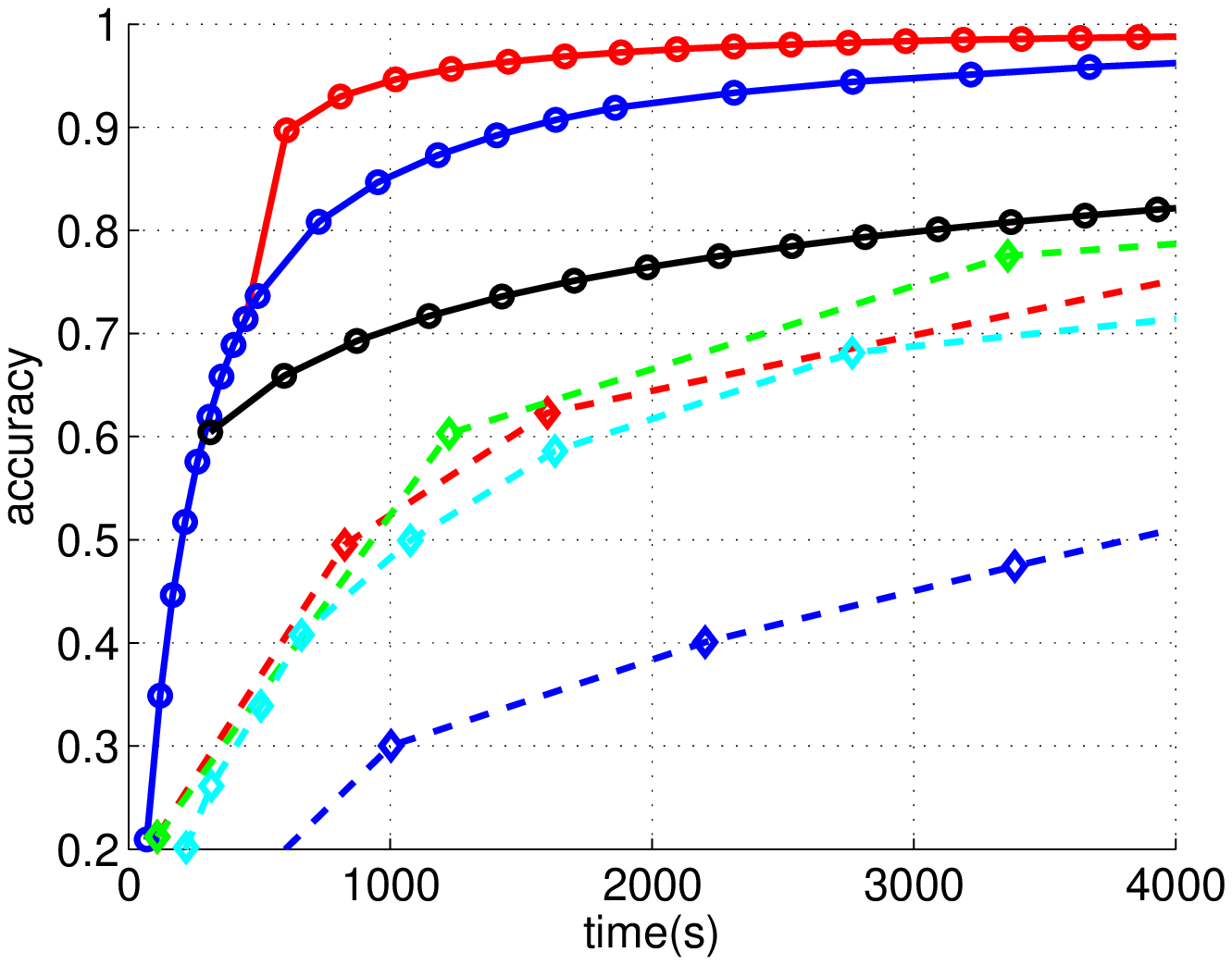}}
\subfigure
{\label{fig:result:uk_20nn}
\includegraphics[width=.2\linewidth, clip]{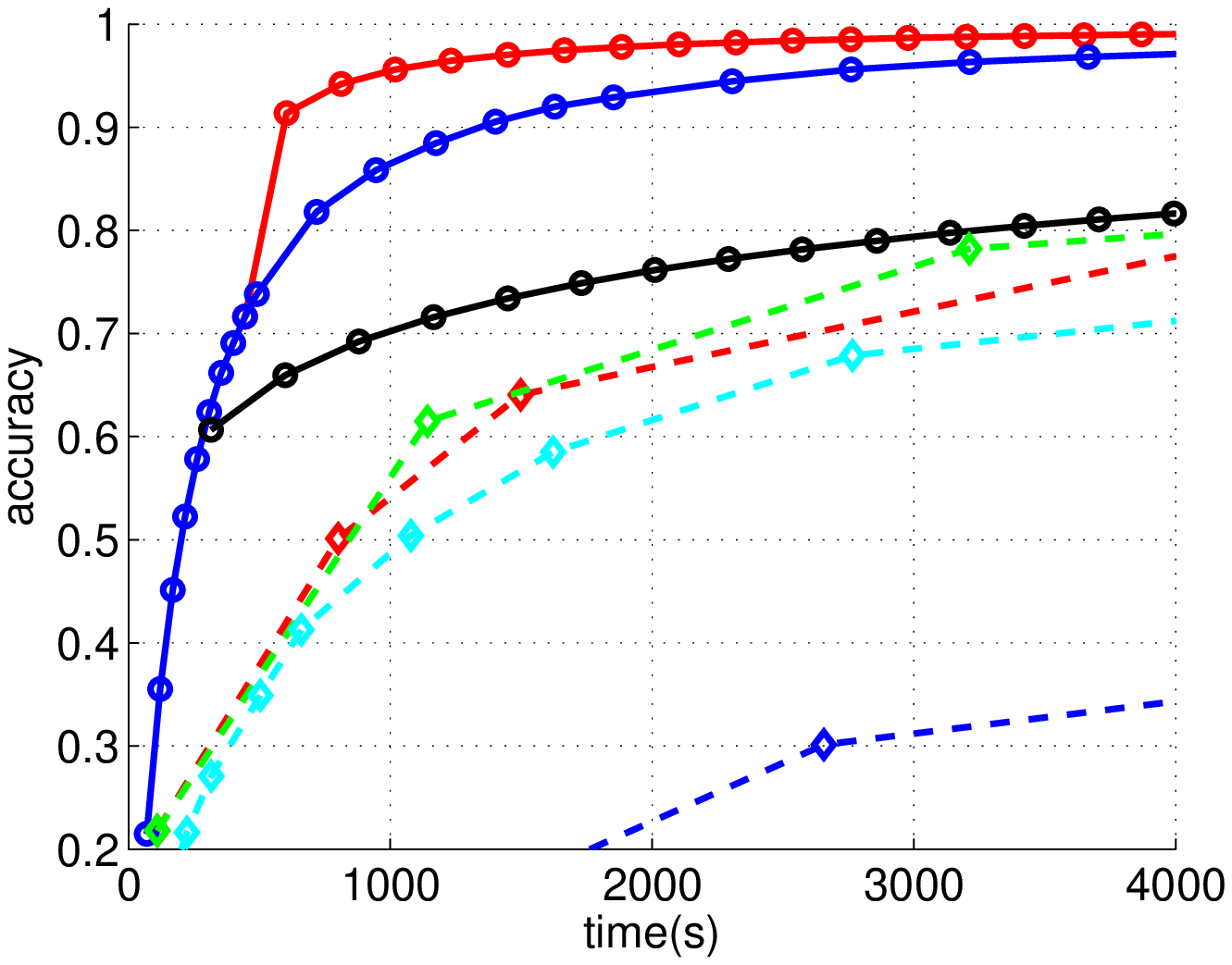}}
\subfigure
{\label{fig:result:imagenet_20nn}
\includegraphics[width=.2\linewidth, clip]{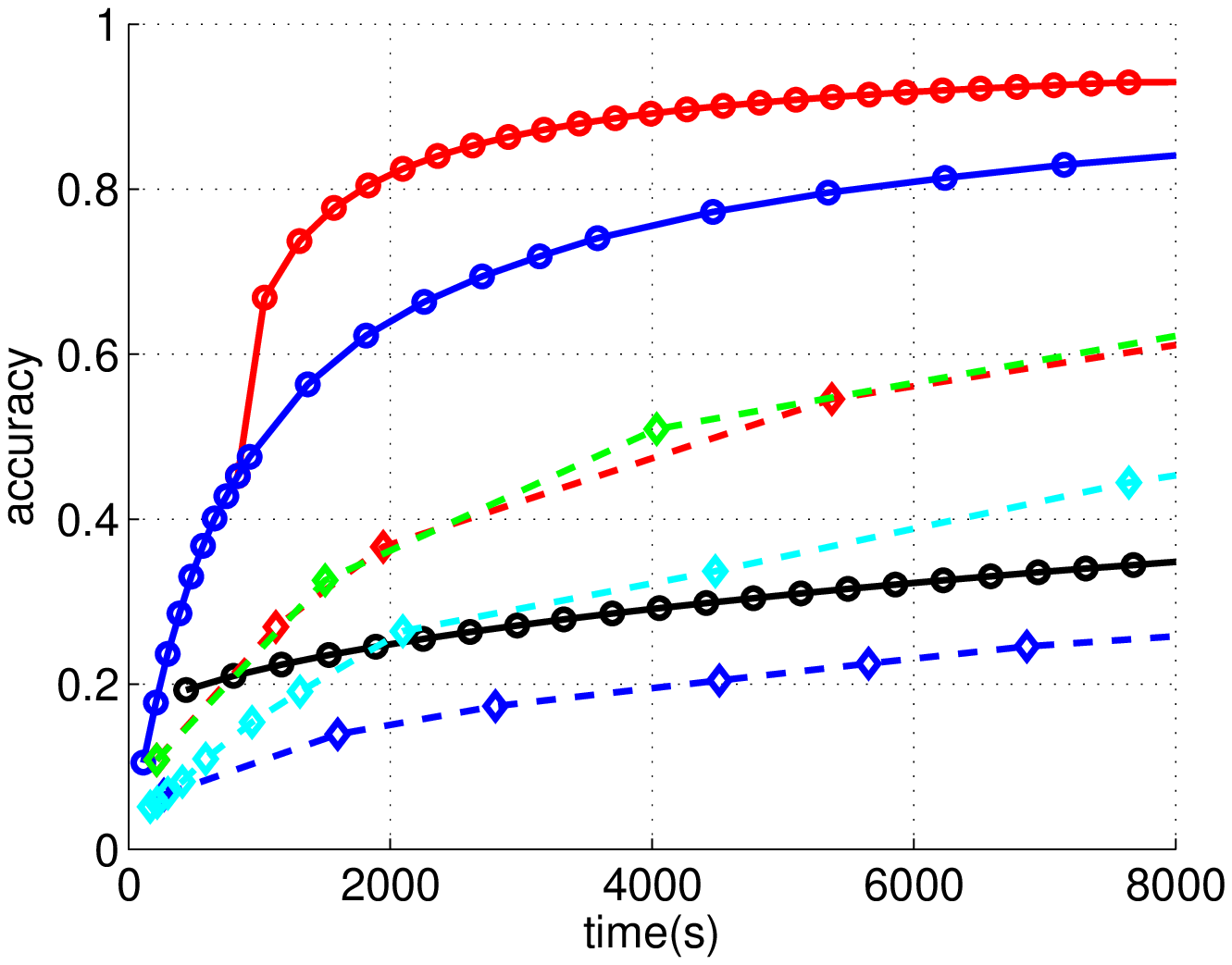}}
\subfigure
{\label{fig:result:tiny_20nn}
\includegraphics[width=.2\linewidth, clip]{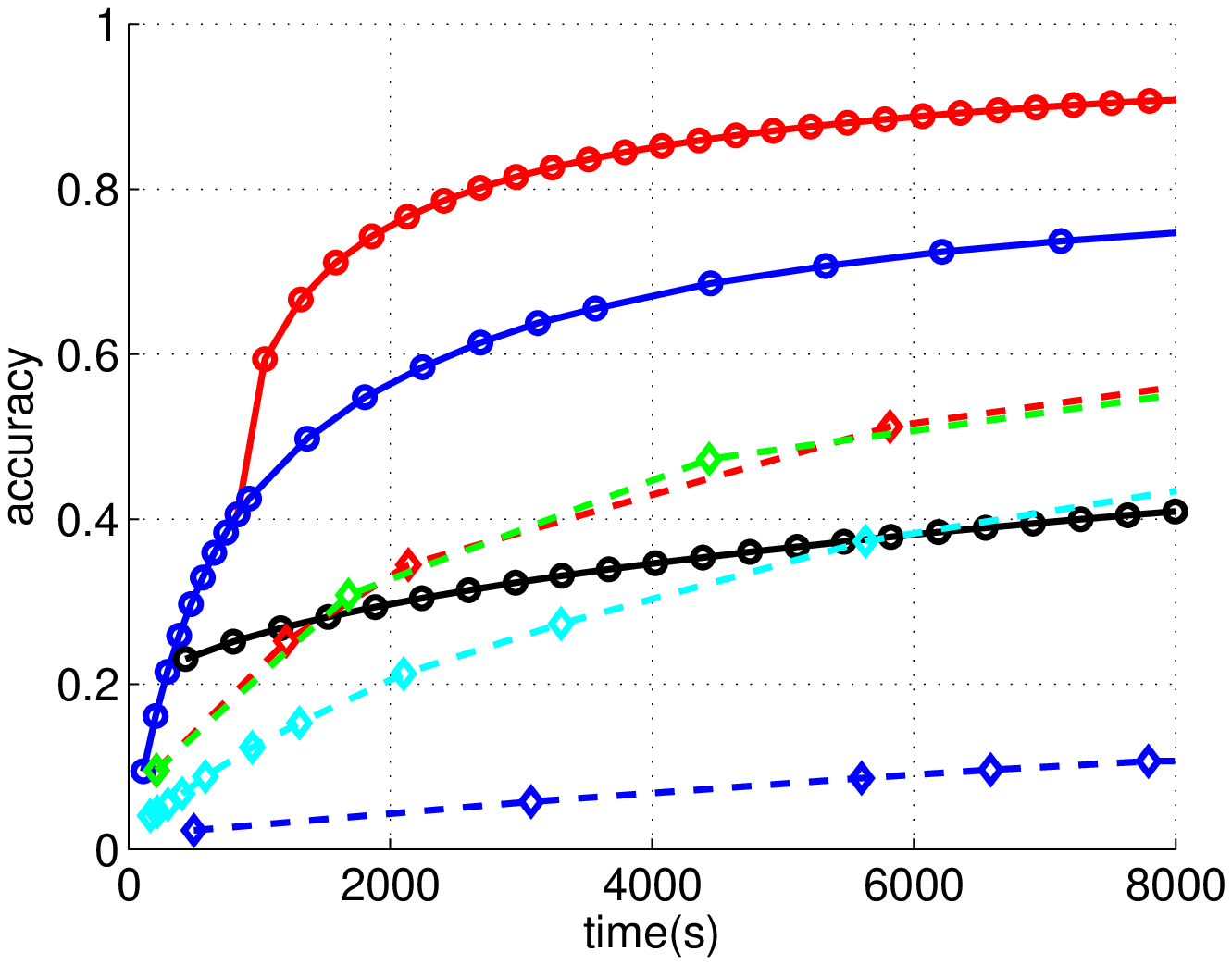}}\\
\scriptsize{$ k = 50 $}
\subfigure[(a)]
{\label{fig:result:101_50nn}
\includegraphics[width=.2\linewidth, clip]{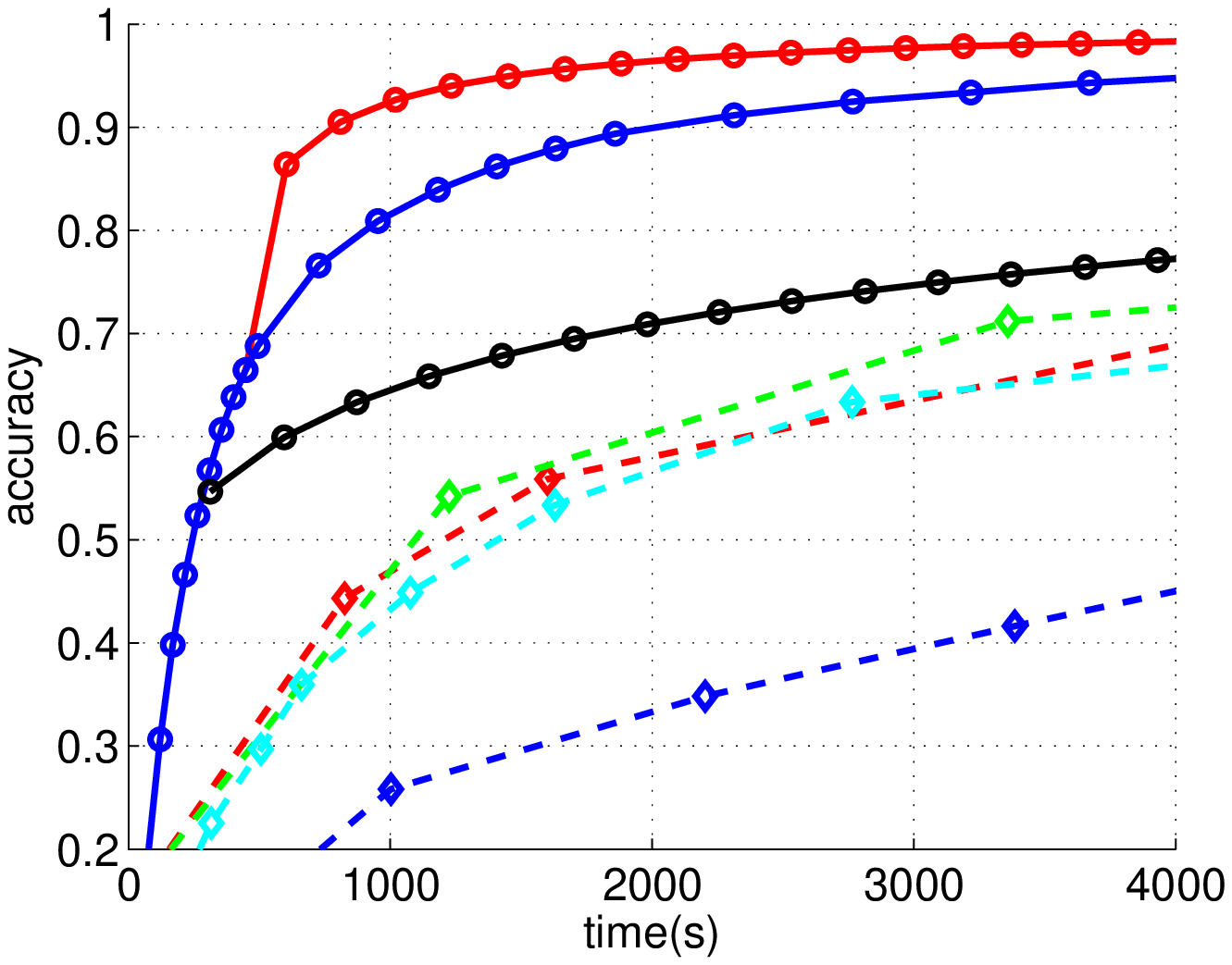}}
\subfigure[(b)]
{\label{fig:result:uk_50nn}
\includegraphics[width=.2\linewidth, clip]{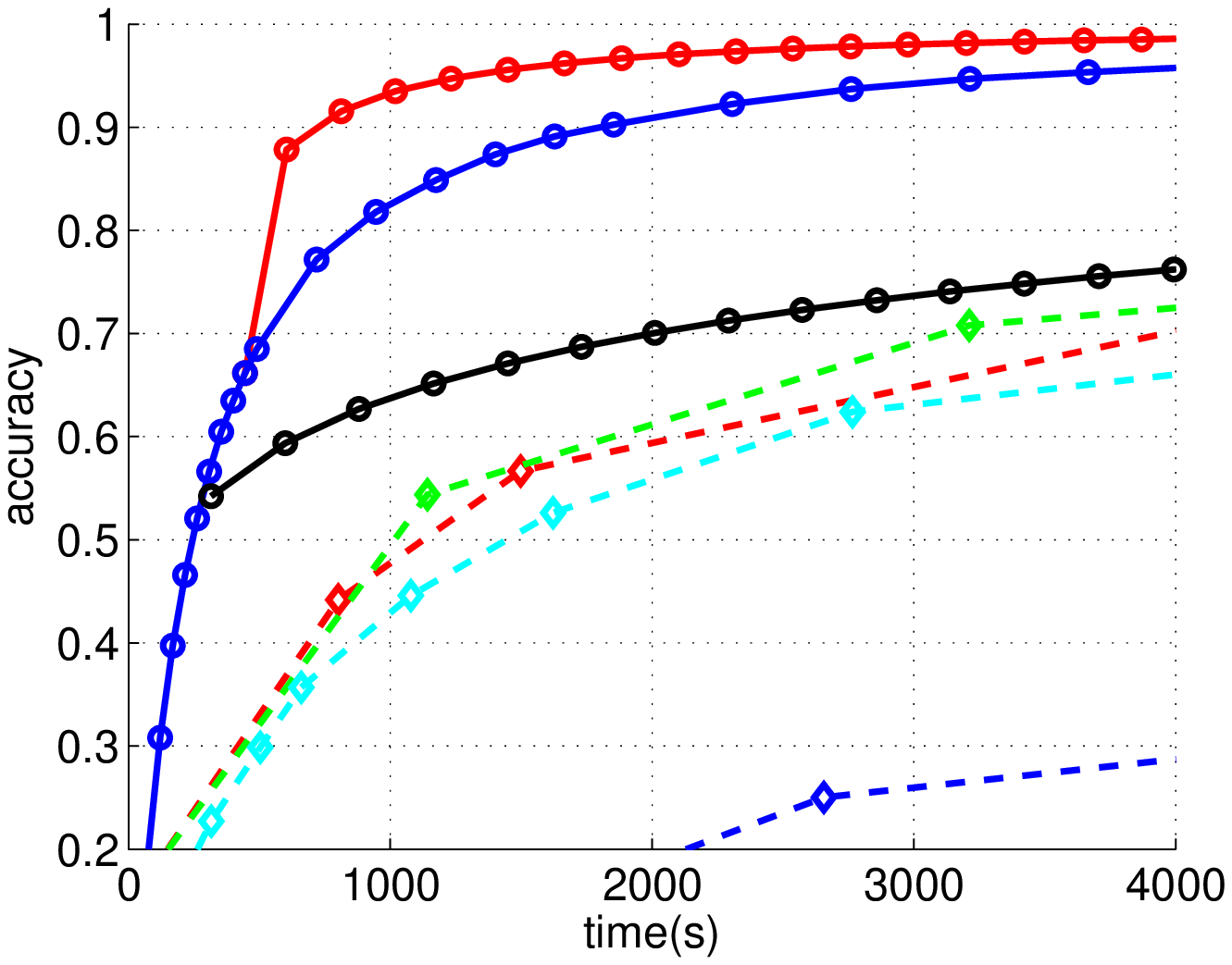}}
\subfigure[(c)]
{\label{fig:result:imagenet_50nn}
\includegraphics[width=.2\linewidth, clip]{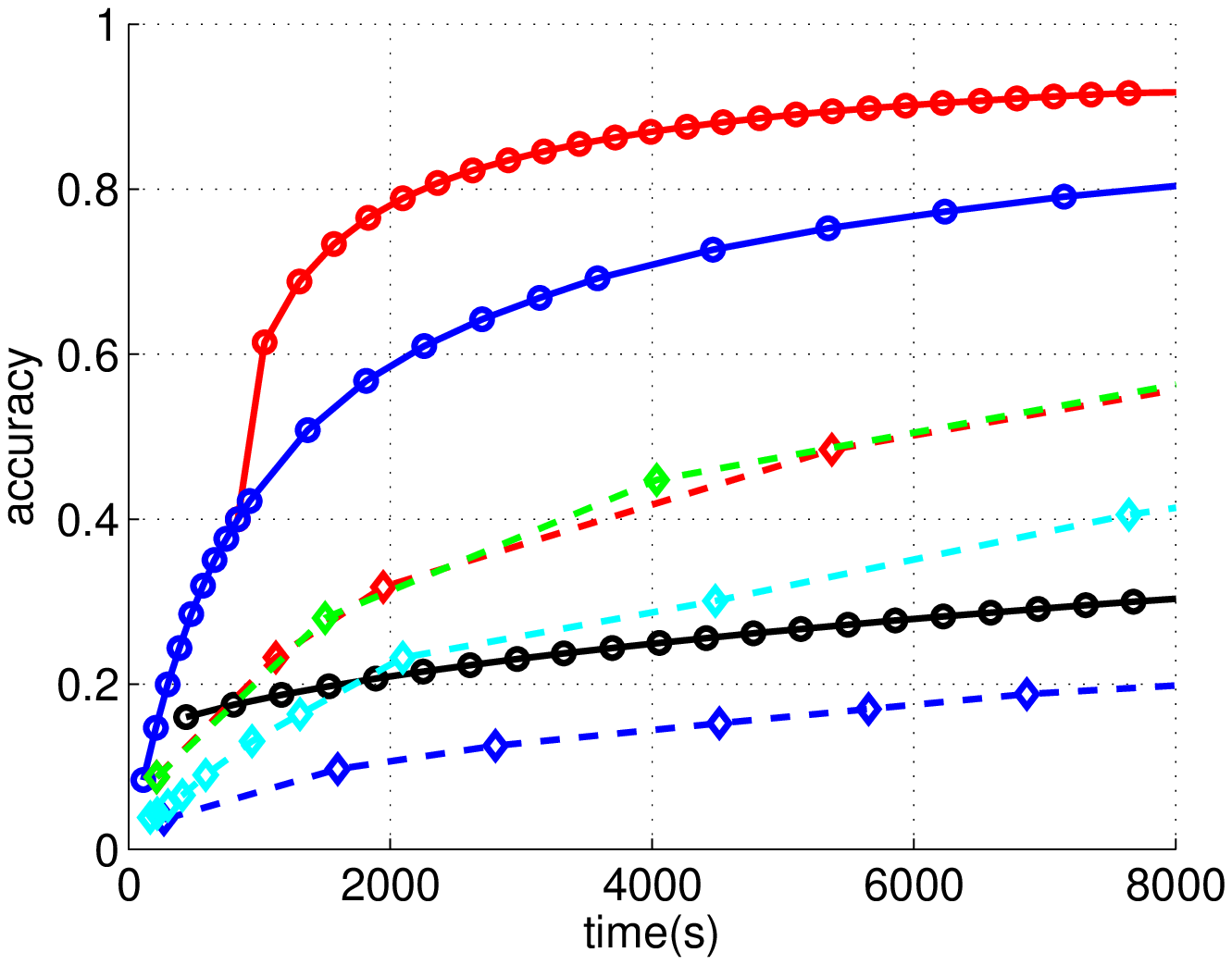}}
\subfigure[(d)]
{\label{fig:result:tiny_50nn}
\includegraphics[width=.2\linewidth, clip]{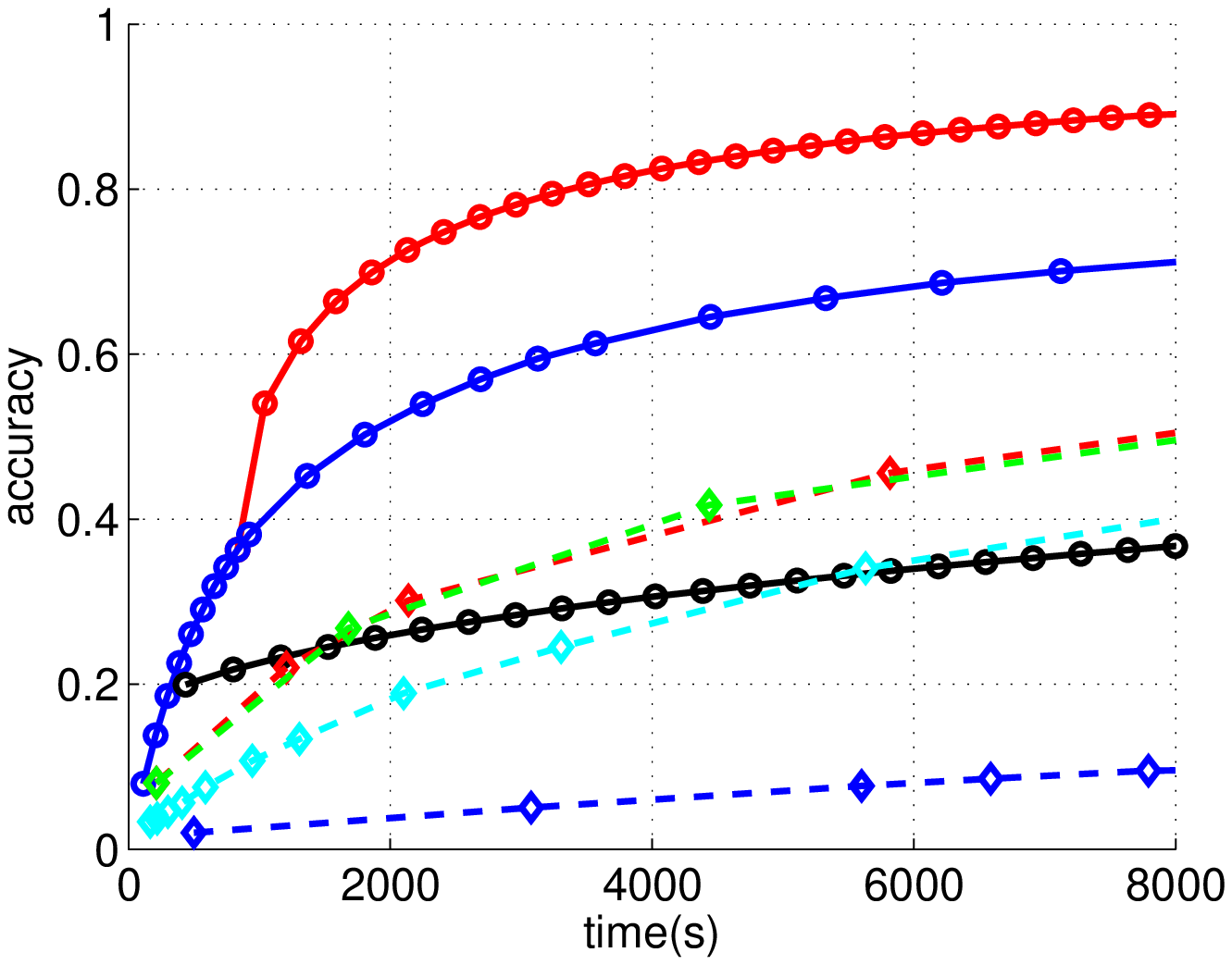}}\\
\label{fig:result}
\centering
\includegraphics[width = 1\linewidth, clip]{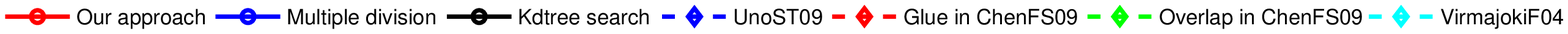}
\MyCaption{Performance comparisons over (a) Caltech 101,
(b) Recognition Benchmark, (c) Imagenet, (d) TinyImage.}
\end{minipage}
\vspace{-0.3cm}
\end{figure*}

\mytextbf{Results.}
The performance comparison is shown in~\Figure{\ref{fig:result}}.
The horizontal axis corresponds to construction time (in seconds),
and the vertical axis corresponds to the accuracy.
We test all the above algorithms
on the four data sets described before,
and compute the accuracy based on
different numbers of neighbors,
denoted as $k$.
Each column of~\Figure{\ref{fig:result}}
corresponds to one data set
and each row corresponds to a choice of $k$.
The performance of multiple random divisions in our approach
is shown as the blue circle line in each figure,
and the performance of the combination of it
with the neighborhood propagation
is shown as the red circle line.

From the results,
we can clearly see the superiority of our algorithms
over other algorithms.
In Caltech 101,
the approach of multiple random divisions
can achieve
an accuracy of $90\%$ in the $1$-NN graph,
at least three times faster than other algorithms,
and when applying neighborhood propagation,
our approach is at least six times faster.
If the targeted accuracy becomes higher,
or the number of neighbors becomes larger,
the improvement becomes more significant.

In the high-dimensional data sets such as TinyImage and Imagenet,
the divide-and-conquer algorithms in~\cite{ChenFS09,VirmajokiF04}
turn out to be more efficient than kd-tree search,
but still at least three times worse than
our multiple random division method
when the required accuracy is above $60\%$.
After adopting neighborhood propagation,
the superiority becomes even more significant
than in the low dimensional cases.
For TinyImage when $k = 50$,
our approach achieves an accuracy of $90\%$
in $6000$ seconds,
but within the same time,
the accuracy of other methods
is at most $50\%$.

Comparing our approach with the brute-force method,
we can see that
our approach achieves $95\%$ in accuracy
using about $1\%$ of the brute-force time
for the $128$-dimensional SIFT features,
and achieves $90\%$ in accuracy
using about $2\%$ of the brute-force time
for the $384$-dimensional GIST features,
which shows that
constructing an approximate neighborhood graph
indeed saves a significant amount of time
with only a minor loss in accuracy.

\section{Applications}

\mytextbf{Face images organization.} We first present an application
that adopts a neighborhood-based distance measure
to organize face images.
The rank-order distance
has been shown good
to evaluate the distance between faces~\cite{ZhuWS11}.
The rank-order distance
over two face images
is computed
by comparing their neighboring faces,
which requires first constructing a $k$-NN face graph.
The data set with about $500K$ face images in our experiment
consists of a labeled face dataset
with $13374$ labeled images from LFW~\cite{LFWTech}
and a distracter face dataset
collected from the Web.
We first compute a $100$-NN graph $\mathcal{G}_1$
using our approach.
Then we conduct the neighborhood propagation step again
to obtain a new $k$-NN graph $\mathcal{G}_2$,
but with the rank-order distances
that are computed using the neighborhood from $\mathcal{G}_1$.
For comparison,
we also report the results over the exact $k$-NN graph $\mathcal{G}_0$
as the baseline.

For evaluation,
we adopt three metrics:
local label consistency ratio,
$\operatorname{Precision}@K$ and $\operatorname{nDCG}@K$.
Local label consistency ratio aims to
evaluate if the label of the face
is dominant among the labels of the $10$ neighboring faces,
and is evaluated as $1$ if the number of faces with the same label is larger than $6$
and otherwise $0$.
$\operatorname{Precision}@K$ is computed as the proportion of the same faces
among the top $K$ neighbors,
and $\operatorname{nDCG}@K$ is the normalized discounted cumulative gain
over the top $K$ neighbors which has been widely used
in various ranking tasks.

\begin{figure}[t]
\centering
\subfigure[(a)]
{
\label{fig:rankorderprecision}
\includegraphics[width=0.45\linewidth]{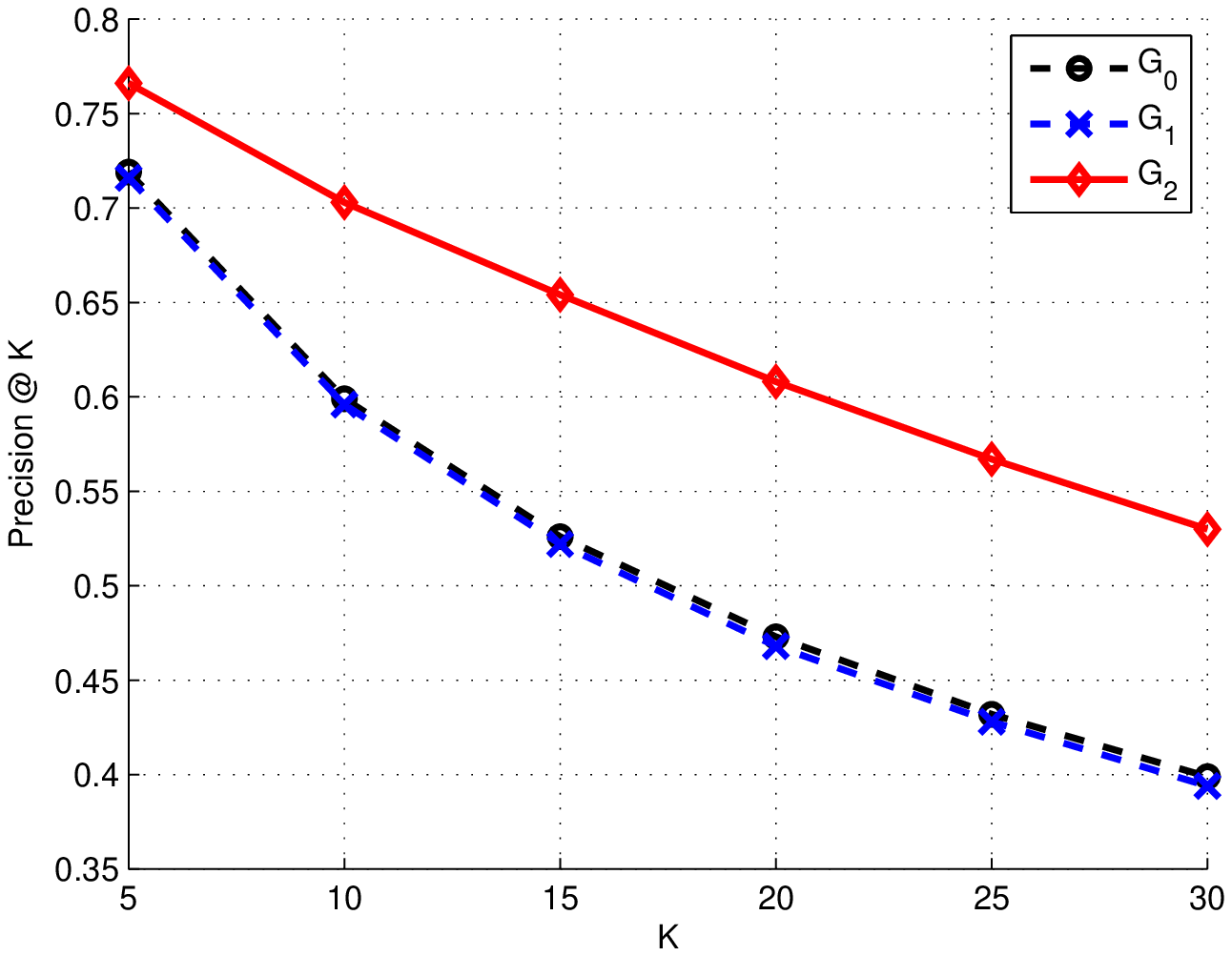}
}
\subfigure[(b)]
{
\label{fig:rankorderndcg}
\includegraphics[width=0.45\linewidth]{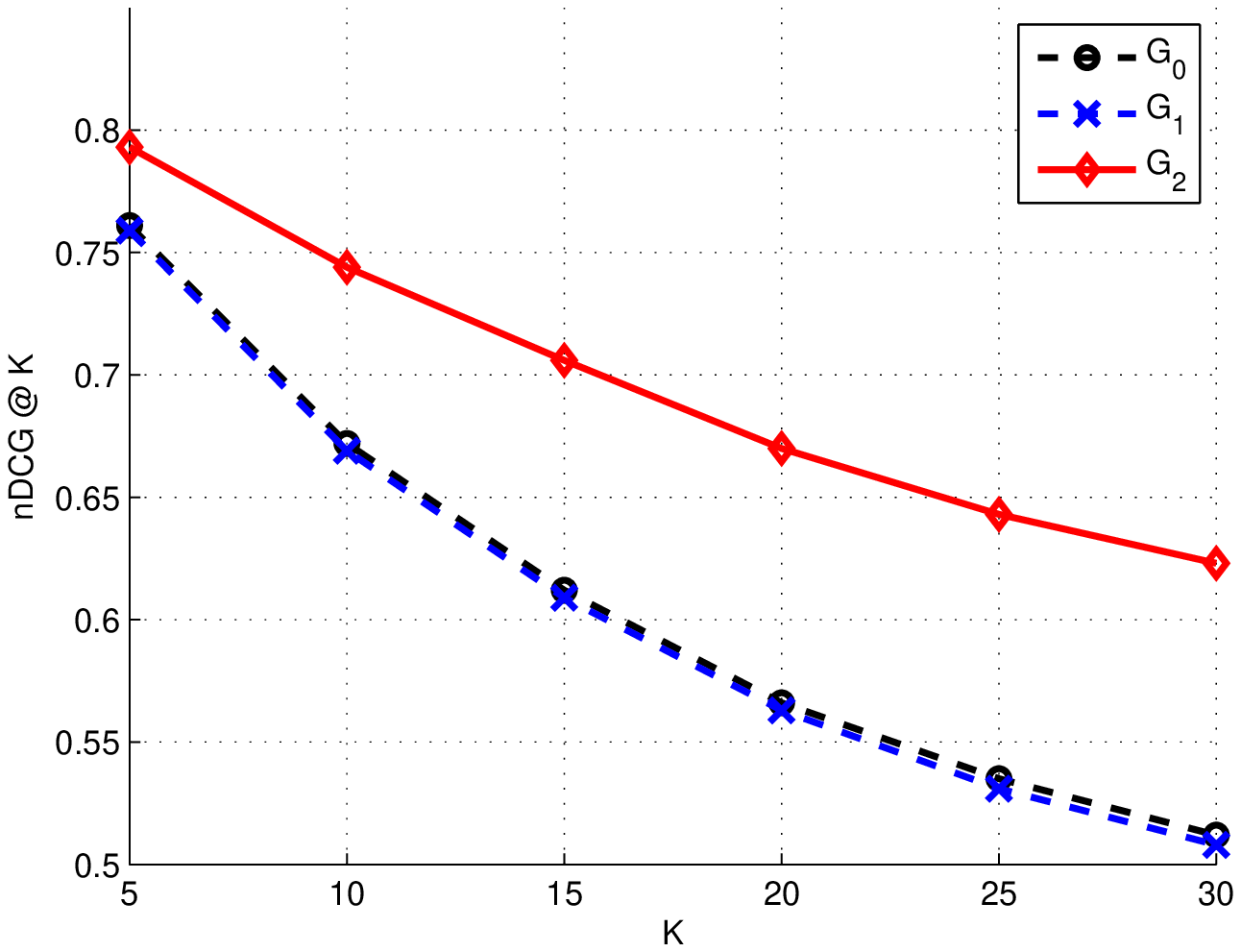}
}
\caption{
Comparison between different ranking results.
(a) $\operatorname{Precision}@K$,
(b) $\operatorname{nDCG}@K$.}
\label{fig:rankorder}
\vspace{-.2cm}
\end{figure}

Tab.~\ref{tab:goodranking} shows the comparison
of the average local label consistency ratio
and Fig.~\ref{fig:rankorder} shows the comparisons
of the average $\operatorname{Precision}@K$ and $\operatorname{nDCG}@K$.
We have two observations.
On the one hand, the neighborhood graph is very powerful and useful,
with which we can get a better distance measure for face images organization.
On the other hand, the performances of $\mathcal{G}_0$ and $\mathcal{G}_1$
are almost the same,
which shows that the $k$-NN graph constructed from our approach
is very accurate.

\mytextbf{Object discovery.}
Discovering objects
from large unlabeled image collections
has been a challenging problem~\cite{ChumM10}.
We show that the proposed approach
can effectively construct a matching graph
with a coarse similarity measure for fast random divisions
and a fine similarity measure for accurate neighborhood propagation
and that objects can be effectively discovered over such a graph.

The data set
consists of $5062$ labeled images from the Oxford $5K$ data set~\cite{PhilbinCISZ07}
and $50K$ distracter images downloaded from Flickr.
SIFT features are extracted from each image.
We build two vocabularies, respectively with $1K$ and $1M$ visual words.
Each image is represented by two features,
one $1K$-dimensional histogram over $1K$ visual words
that indicates word occurrences
and is then weighted by tf-idf,
and one bag-of-words representation over $1M$ visual words
with attaching the spatial position for each word.
The low dimensional features are used to build random divisions
for fast neighborhood graph construction,
and the high dimensional features with its spatial information
are then used to compute image matching
with spatial verification~\cite{PhilbinCISZ07}
for accurate neighborhood propagation,
yielding a matching graph.
We run affinity propagation~\cite{FreyD07}
over this matching graph to
over-segment the image set
so that similar views of the same object
are grouped together.
To join these over-segments
for grouping images with the same object,
we then construct a graph with over-segments as nodes,
and define the similarity between two over-segments
as the proportion of images that in both over-segments are
$50$-reciprocal nearest neighbors~\cite{QinGBQG11},
which is fast computed over the matching graph.
We finally apply affinity propagation again
over the over-segment graph to obtain the final clustering result.

To evaluate the performance,
for each labeled object group,
we find the cluster containing the most images of that object
and compute precision, recall and F-measure of that cluster.
The result is given in Tab.~\ref{tab:oxford}.
We can see that most objects can be discovered
with an F-measure over $0.6$.

\begin{table}[t]
\MyCaption{Performance for object discovery.}
\label{tab:oxford}
\centering
{\footnotesize
{
\begin{tabular}{c|c||c|c|c}
\hline
& \#images & Precision & Recall & F-measure \\
\hline
All Souls & 78 & 0.785 & 0.654 & 0.713 \\
Ashmolean & 25 & 0.947 & 0.720 & 0.818 \\
Balliol & 12 & 0.800 & 0.333 & 0.471 \\
Bodleian & 24 & 0.100 & 0.542 & 0.169 \\
Christ Church & 78 & 0.360 & 0.692 & 0.474 \\
Cornmarket & 9 & 0.833 & 0.556 & 0.667 \\
Hertford & 54 & 0.829 & 0.630 & 0.716 \\
Keble & 7 & 0.667 & 0.571 & 0.615 \\
Magdalen & 54 & 1.000 & 0.130 & 0.230 \\
Pitt Rivers & 6 & 1.000 & 0.833 & 0.909 \\
Radcliffe Camera & 221 & 0.820 & 0.661 & 0.732 \\
\hline
Average & & 0.740 & 0.574 & 0.592 \\
\hline
\end{tabular}}
\vspace{-.2cm}
}
\end{table}

\section{Conclusions}
In this paper, we address the problem
of constructing $k$-NN graphs
for large scale visual descriptors.
Our approach consists of
two steps: multiple random divide-and-conquer
and neighborhood propagation.
We show both theoretical and empirical
accuracy and efficiency
of our approach.
As an ongoing work,
we are investigating a learning scheme
to automatically trigger neighborhood propagation.

%

\section*{Appendix}
\label{sec:appendix}
\subsection*{Proofs}

\begin{Lemma}
\label{lemma1}
Suppose a random hyperplane partitions the data points
so that a certain point $\mathbf{x}_i$
and one of its true neighbors $\mathbf{x}_j$
have the probability $P_{ij}$ to be on the same side.
Then with a single random partition tree,
$\mathbf{x}_j$ is discovered as $\mathbf{x}_i$'s neighbor
with the probability $P_{ij}^h$,
where $h$ is the depth of the tree.
With $L$ random partitions,
$\mathbf{x}_j$ is discovered as $\mathbf{x}_i$'s neighbor
with the probability $1 - (1 - P_{ij}^h)^L$.
\end{Lemma}

\begin{proof}
In a random partition tree,
two data points $\mathbf{x}_i$ and $\mathbf{x}_j$
will discover each other as neighbors if they
they lie on the same side of $h$ random hyperplanes,
where $h$ is the height of the tree.
According to the assumption,
for each hyperplane $\mathbf{x}_i$ and $\mathbf{x}_j$
will lie one the same side with probability $P_{ij}$.
Because hyperplanes are independent with each other,
$\mathbf{x}_i$ and $\mathbf{x}_j$ will
discover each other as a neighbor
with the probability $P_{ij}^h$.
Moreover, the partition trees are
also independent with each other,
so that with $L$ trees,
the discovery probability will be $1-(1-P_{ij}^h)^L$.
\end{proof}

This property can easily be validated
using the similar manner to LSH~\cite{DatarIIM04}.
Although there is a slight difference between random partition trees
and LSH that a random partition tree for each level
may use different projections,
the statement that a true neighboring point of $\mathbf{x}$ is discovered
with the probability $P_{ij}^h$
still holds
because discovering a true neighboring point of $\mathbf{x}$
must pass $h$ projections.
In the case of Euclidean distance,
the stable distribution~\cite{DatarIIM04} can be used to
build a binary partition
and the probability $P_{ij}$ can be also easily computed.
For example,
if using $h_{\mathbf{a}, b}(\mathbf{x}) = \lfloor \frac{\mathbf{a}^T \mathbf{x} + b}{w}\rfloor$
where each entry of $\mathbf{a}$ satisfies a Gaussian distribution
and $b$ is a uniform random variable over $[0, w]$.
It can be demonstrated that
$P(h_{\mathbf{a},b}(\mathbf{x}_i) = h_{\mathbf{a},b}(\mathbf{x}_j))
= \int_{0}^{w} \frac{1}{d_{ij}}f_p(\frac{t}{d_{ij}})(1 - \frac{t}{w})dt$,
where $d_{ij}$ is the Euclidean distance between $\mathbf{x}_i$ and $\mathbf{x}_j$
and $f_p(t)$ denotes the probability density function of the absolute value of
Gaussian distribution.
The probability will be larger than $\frac{1}{2}$
when $w \geqslant 2d_{ij}$.
Then we can bi-partition the data
according to the median of $h_{\mathbf{a}, b}(\mathbf{x})$
so that $P_{ij}$ will be larger than $\frac{1}{2}$
when $w \geqslant 2d_{ij}$.
In the case of cosine distance~\cite{DatarIIM04},
a random projection can be used to build the random partition tree
and $P_{ij}$ can be computed as $1 - \frac{d_{ij}}{\pi}$.

\begin{Lemma}
\label{lemma2}
The probability that
the neighboring relationship between $\mathbf{x}_i$ and $\mathbf{x}_j$ is discovered
by the $L$-th random partition tree
but not discovered by the previous $L-1$ random partition trees
is $ P_{r_L} = (1 - P_{ij}^h)^{L-1} P_{ij}^h$.
\end{Lemma}

\begin{proof}
According to \LemmaRef{\ref{lemma1}},
for each of the previous $L-1$ random partition trees,
the probability that
$\mathbf{x}_i$ and $\mathbf{x}_j$
fail to discover each other as neighbors is $1-P_{ij}^h$,
and in the $L$-th tree the neighboring relationship
will be discovered with the probability $P_{ij}^h$.
Then the lemma can be proved with the basic multiplication principle.
\end{proof}

\begin{Lemma}
\label{lemma3}
Considering two neighboring points $\mathbf{x}_i$ and $\mathbf{x}_j$
having the same neighboring point $\mathbf{x}_n$,
after $L-1$ partition trees,
the probability that $\mathbf{x}_j$ can be discovered as the neighbor of $\mathbf{x}_i$
through $\mathbf{x}_n$
is $ P_{p_L} = (1 - (1- P_{in}^h)^{L-1})(1 - (1- P_{jn}^h)^{L-1})$.
\end{Lemma}

\begin{proof}
$\mathbf{x}_j$ can be discovered as the neighbor of $\mathbf{x}_i$
through $\mathbf{x}_n$ if and only if
the neighboring relationship between $\mathbf{x}_i$ and $\mathbf{x}_n$
and the relationship between $\mathbf{x}_j$ and $\mathbf{x}_n$
have both been discovered.
From \LemmaRef{\ref{lemma1}}, we know the probability of the two events
are $(1 - (1- P_{in}^h)^{L-1})$ and $(1 - (1- P_{jn}^h)^{L-1})$,
by simply multiplying them,
we get $ P_{p_L} = (1 - (1- P_{in}^h)^{L-1})(1 - (1- P_{jn}^h)^{L-1})$.
\end{proof}

\begin{Theorem}
\label{theorem1}
Suppose $P = \min\{ P_{ij} | \langle \mathbf{x}_i,\mathbf{x}_j \rangle \in E(G) \}$,
where $G$ is the exact $k$-NN graph of the data set.
With $L$ random divisions and a first-order neighborhood propagation
a true $k$-NN point of $\mathbf{x}_i$, $\mathbf{x}_j$ is discovered
with at least the probability $1 - (1 - P^h)^{2L}(2-(1-P^h)^L)$
under the assumption
that $\mathbf{x}_i$ and $\mathbf{x}_j$
have at least one same neighboring point $\mathbf{x}_n$.
\end{Theorem}

\begin{proof}
With two ways $\mathbf{x}_i$ will find $\mathbf{x}_j$
as a neighbor.
First is to discover $\mathbf{x}_j$ in at least one of the $L$ partition trees,
and the probability of it is $1 - (1 - P_{ij}^h)^L$ according to \LemmaRef{\ref{lemma1}}.
Second is fail to discover $\mathbf{x}_j$ in partition trees,
but discover it during the first-order neighborhood propagation.
From \LemmaRef{\ref{lemma2}},\ref{lemma3} the probability is
$(1-P_{ij}^h)^L(1 - (1- P_{in}^h)^{L})(1 - (1- P_{jn}^h)^{L})$.
To sum them up, the probability that
$\mathbf{x}_i$ discovers $\mathbf{x}_j$
with $L$ partition trees and a first-order neighborhood propagation is
\begin{align}
\centering
& ~1 - (1 - P_{ij}^h)^L \nonumber + (1-P_{ij}^h)^L(1 - (1- P_{in}^h)^{L})(1 - (1- P_{jn}^h)^{L})
\nonumber \\
\geqslant & ~1 - (1 - P_{ij}^h)^L + (1-P_{ij}^h)^L(1 - (1- P^h)^L)^2 \nonumber \\
&~(P_{in},P_{jn} \geqslant P)
\nonumber \\
= & ~1 - (1 - P_{ij}^h)^L + (1-P_{ij}^h)^L( 1 - 2(1-P^h)^L + (1-P^h)^{2L} ) \nonumber \\
= & ~1 - (1 - P_{ij}^h)^L ( 2(1-P^h)^L - (1-P^h)^{2L} ) \\
\because~&~0 < P < 1 \nonumber \\
\therefore~&~2(1-P^h)^L - (1-P^h)^{2L} > 0 \nonumber \\
(1) > & ~1 - (1-P^h)^L( 2(1-P^h)^L - (1-P^h)^{2L} ) \nonumber \\
= & ~1 - (1 - P^h)^{2L}(2 - (1-P^h)^L) \nonumber
\end{align}
\end{proof}


\subsection*{Face image organization}

\begin{figure*}
\begin{minipage}{1\linewidth}
\centering
\subfigure
{\label{1_good}
\includegraphics[width = 0.49\linewidth, clip]{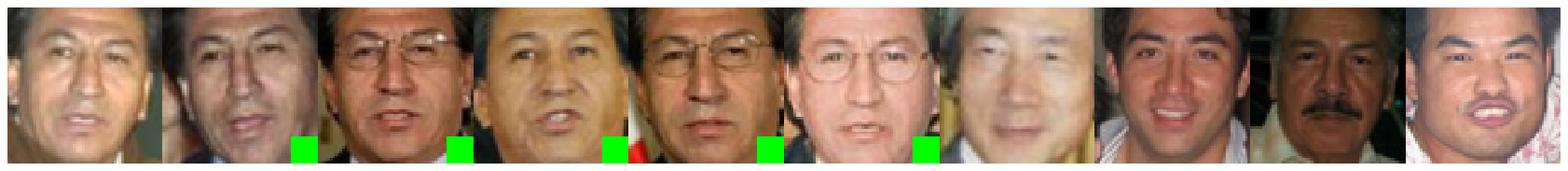}}
\subfigure
{\label{1_bad}
\includegraphics[width = 0.49\linewidth, clip]{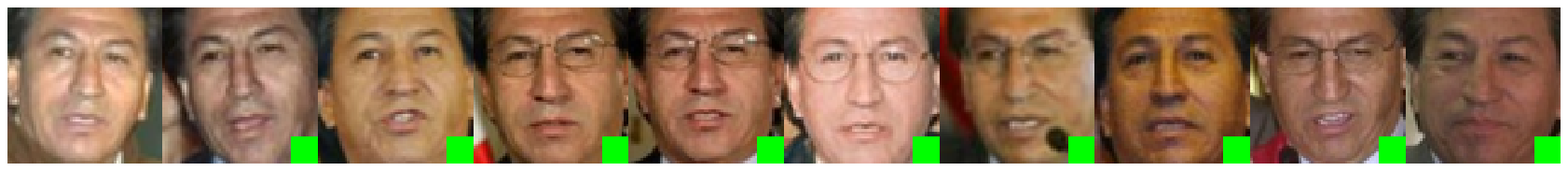}}\\
\vspace{-.1cm}
\subfigure
{\label{2_good}
\includegraphics[width = 0.49\linewidth, clip]{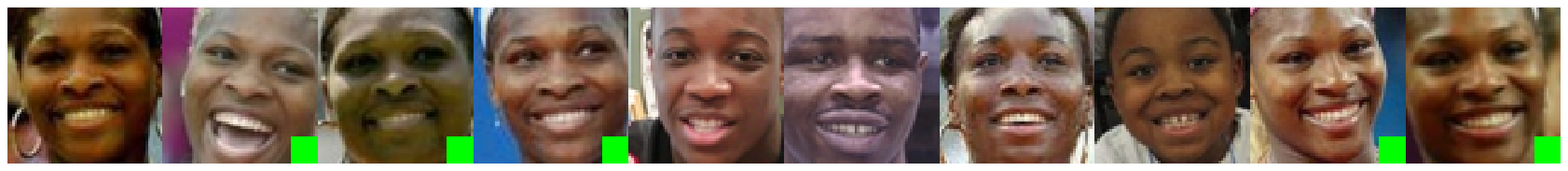}}
\subfigure
{\label{2_bad}
\includegraphics[width = 0.49\linewidth, clip]{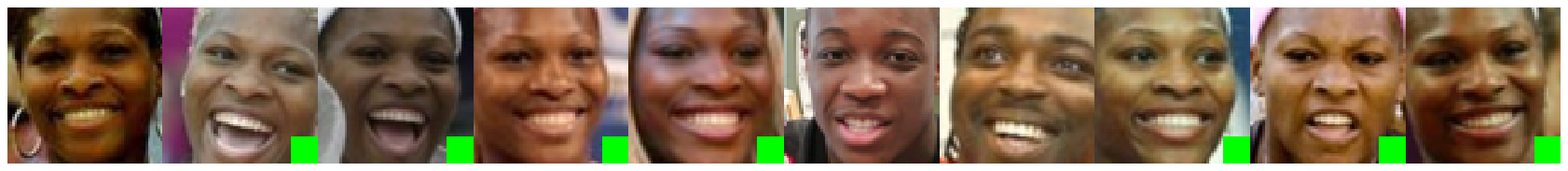}}\\
\vspace{-.1cm}
\subfigure
{\label{3_good}
\includegraphics[width = 0.49\linewidth, clip]{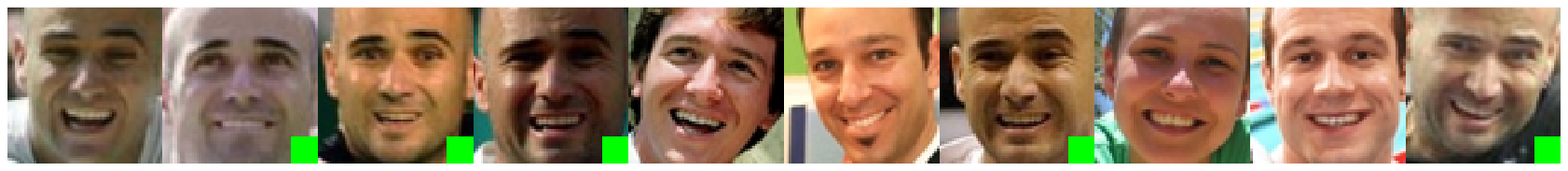}}
\subfigure
{\label{3_bad}
\includegraphics[width = 0.49\linewidth, clip]{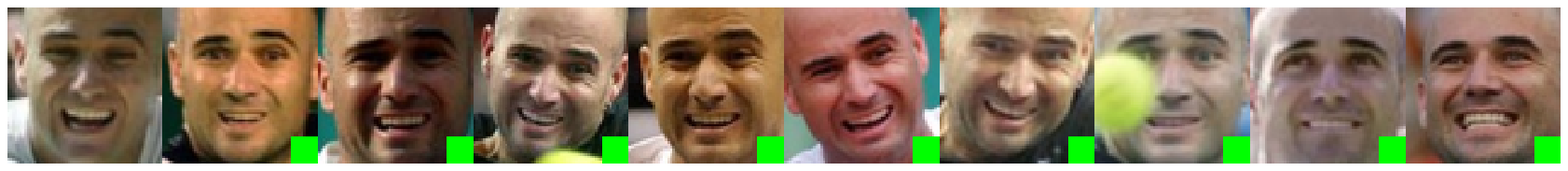}}\\
\vspace{-.1cm}
\subfigure
{\label{4_good}
\includegraphics[width = 0.49\linewidth, clip]{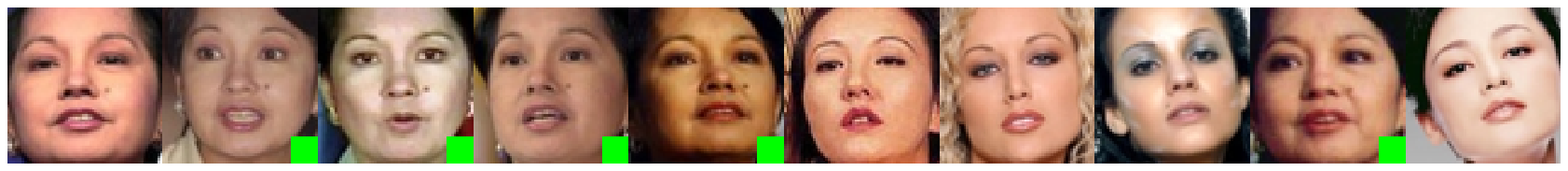}}
\subfigure
{\label{4_bad}
\includegraphics[width = 0.49\linewidth, clip]{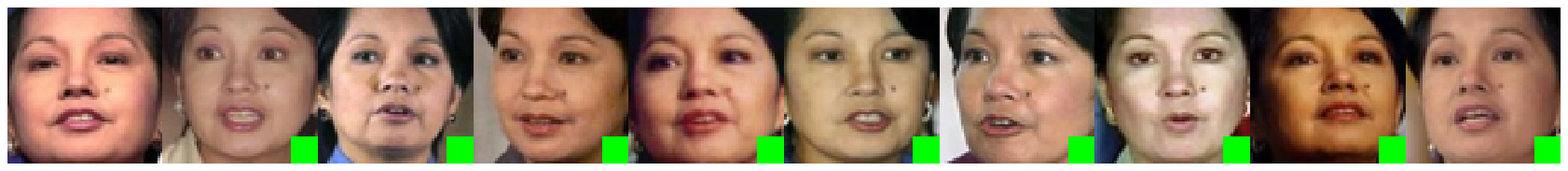}}\\
\vspace{-.1cm}
\subfigure
{\label{5_good}
\includegraphics[width = 0.49\linewidth, clip]{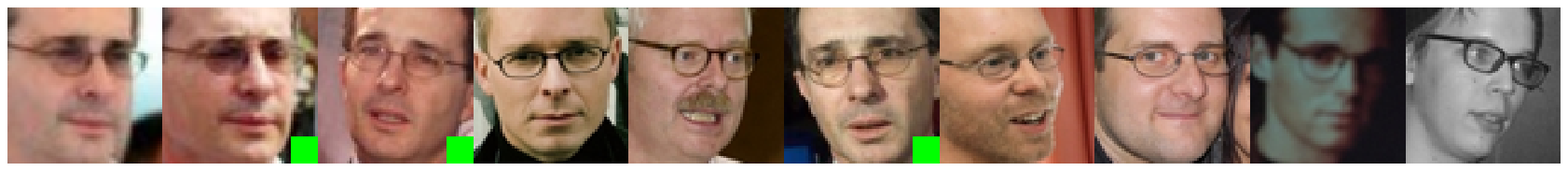}}
\subfigure
{\label{5_bad}
\includegraphics[width = 0.49\linewidth, clip]{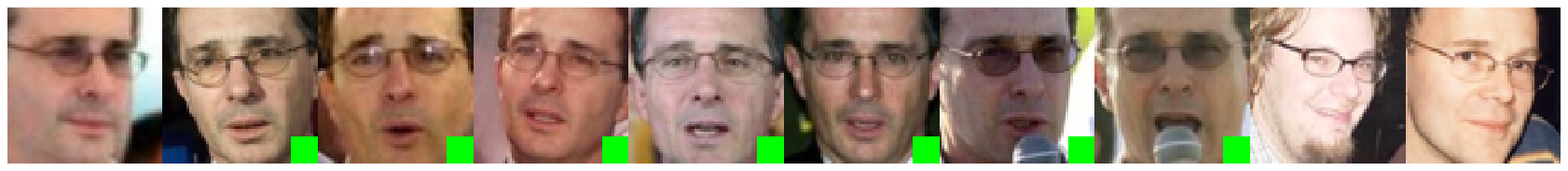}}\\
\vspace{-.1cm}
\subfigure
{\label{6_good}
\includegraphics[width = 0.49\linewidth, clip]{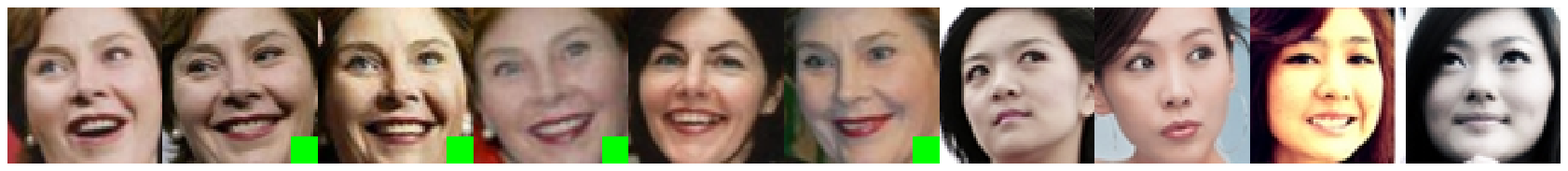}}
\subfigure
{\label{6_bad}
\includegraphics[width = 0.49\linewidth, clip]{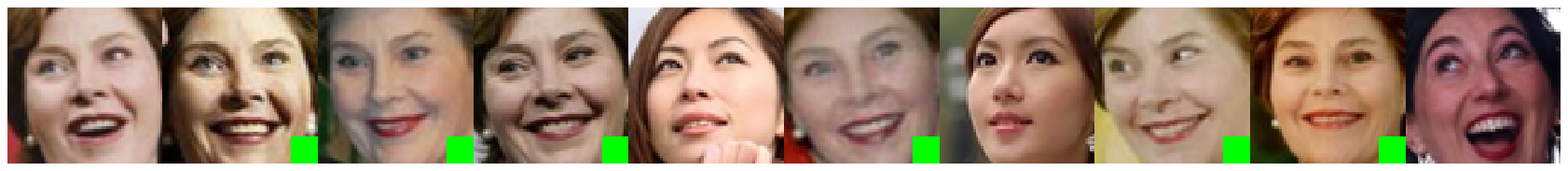}}\\
\vspace{-.1cm}
\subfigure[(a)]
{\label{7_good}
\includegraphics[width = 0.49\linewidth, clip]{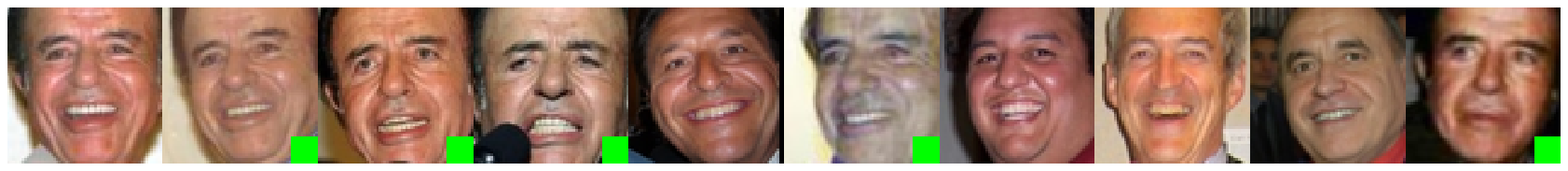}}
\subfigure[(b)]
{\label{7_bad}
\includegraphics[width = 0.49\linewidth, clip]{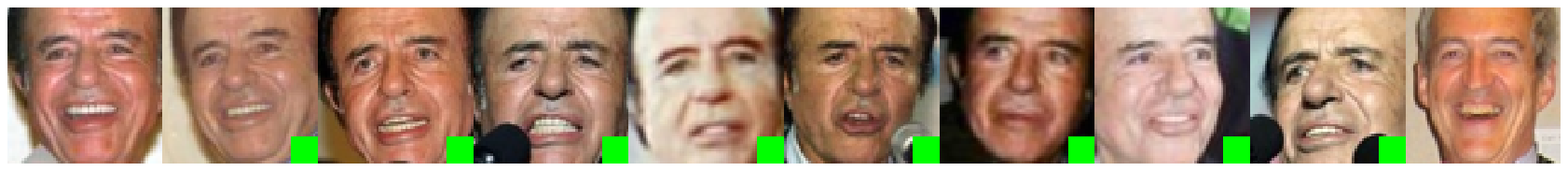}}\\
\MyCaption{Image ranking result comparison between
rank-order distance and Euclidean distance.
The first image in each row are the query
and the images with green dot at bottom right
are the correct images containing the same face as the query.
(a) gives the results of using Euclidean distance
and (b) gives the results of using Rank-order distance.
}
\label{fig:RankOrderVisual}
\end{minipage}
\end{figure*}

Fig.~\ref{fig:RankOrderVisual} shows some examples
of face organization results
when using Euclidean distance and Rank-order distance.
For each face image,
we show its 9 nearest neighbors in the graph,
and it can be clearly seen that with Rank-order distance,
the label consistency within each neighborhoods is enhanced.

\subsection*{Object Discovery}

\begin{figure}
\centering
\subfigure
{\label{oxford_1}
\includegraphics[width = 1\linewidth, clip]{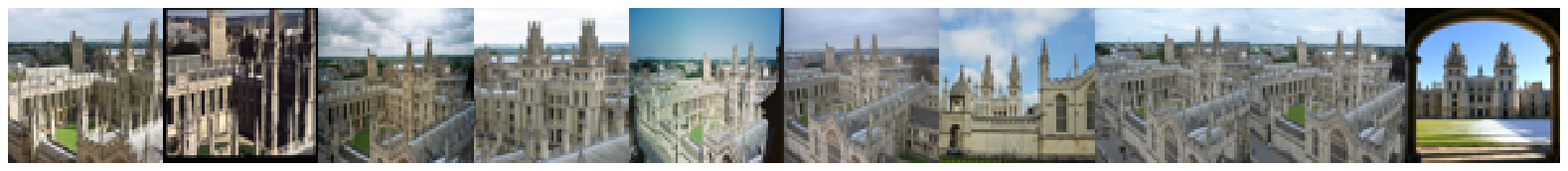}}\\
\vspace{-.1cm}
\subfigure
{\label{oxford_2}
\includegraphics[width = 1\linewidth, clip]{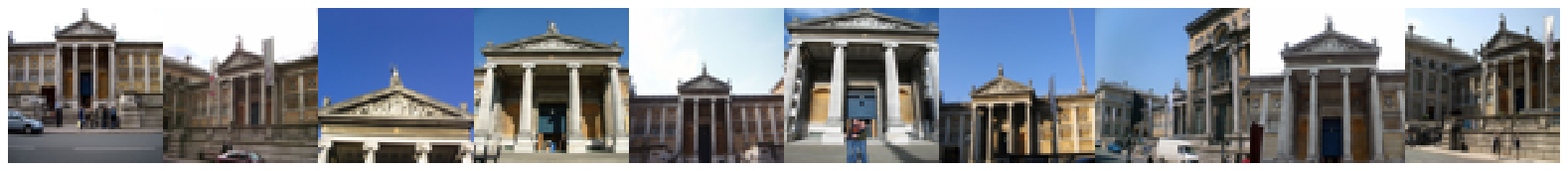}}\\
\vspace{-.1cm}
\subfigure
{\label{oxford_3}
\includegraphics[width = 1\linewidth, clip]{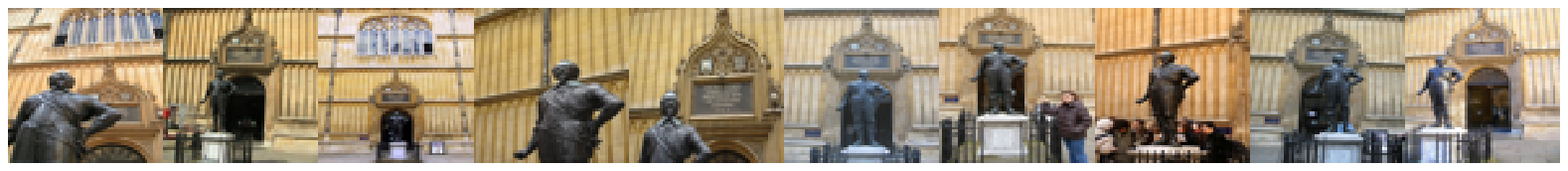}}\\
\vspace{-.1cm}
\subfigure
{\label{oxford_4}
\includegraphics[width = 1\linewidth, clip]{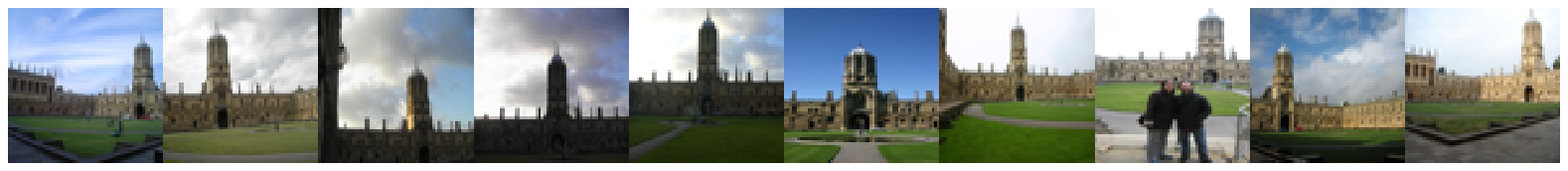}}\\
\vspace{-.1cm}
\subfigure
{\label{oxford_5}
\includegraphics[width = 1\linewidth, clip]{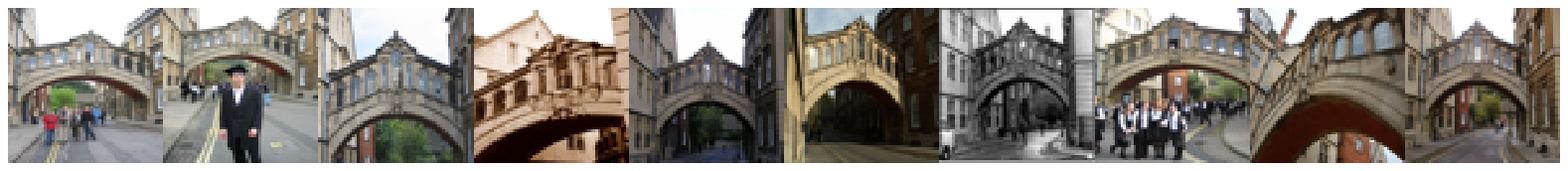}}\\
\vspace{-.1cm}
\subfigure
{\label{oxford_6}
\includegraphics[width = 1\linewidth, clip]{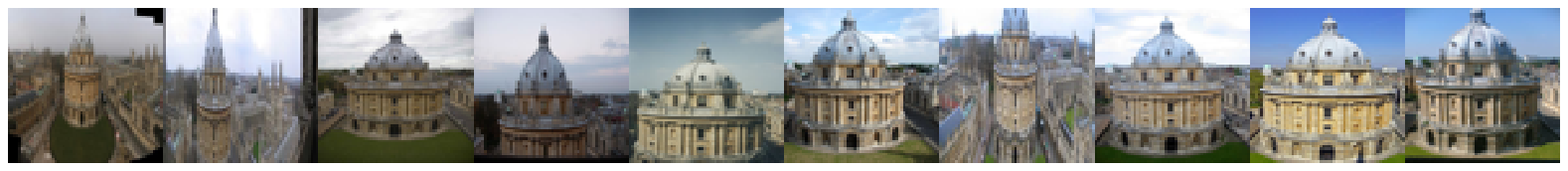}}\\
\MyCaption{Object discovery results:
each row represents one discovered object cluster
and the images are randomly picked out from the cluster.
}
\label{fig:ObjectDiscoveryVisual}
\end{figure}

Fig.~\ref{fig:ObjectDiscoveryVisual} shows some examples
of the detected clusters in the experiments
of object discovery.

{
\footnotesize
\bibliographystyle{ieee}
\bibliography{bow}
}

\end{document}